\documentclass[a4paper, 12pt]{article}
\usepackage{graphicx}
\usepackage[left=2cm, right=2cm, top=2cm, bottom=2cm]{geometry}
\usepackage{amsfonts}
\usepackage{amsmath}
\usepackage{mathrsfs} % needed for \mathscr
\usepackage{color}
\usepackage[numbers,sort&compress]{natbib}
\usepackage{hyperref}
\usepackage[toc,page]{appendix}
\usepackage{authblk}
\usepackage[per-mode=symbol]{siunitx}
\usepackage[percent]{overpic}
\usepackage{multirow}
\usepackage{pgfplots}
\usepackage{tikz}
\usepackage{transparent}
\usepackage{nth}
\usetikzlibrary{shapes,arrows}

\hypersetup{linkcolor=blue, colorlinks =true, citecolor=blue, urlcolor=magenta} 

\DeclareSIUnit\PSI{psi}
\newcommand{\II}{\mathbf{I}}
\newcommand{\uu}{\mathbf{u}}
\newcommand{\F}{\mathbf{F}}
\newcommand{\qq}{\mathbf{q}}
\newcommand{\XX}{\mathbf{X}}
\newcommand{\LLambda}{\mathbf{\Lambda}}
\newcommand{\ddelta}{\boldsymbol{\delta}}
\newcommand{\dLambda}{\delta \mkern-1mu \Lambda}
\newcommand{\ddLambda}{\boldsymbol{\delta}\mkern-2mu\mathbf{\Lambda}}
\newcommand{\TT}{\mathbf{T}}
\newcommand{\RR}{\mathcal{R}}

\newcommand{\Norm}[1]{\left\lVert#1\right\rVert}
\newcommand{\Det}[1]{\left| #1 \right|}
\newcommand{\Dev}[1]{\text{dev}\!\left(#1\right)}

\newcommand{\detF}{\Det{\F}}

\makeatletter
\DeclareRobustCommand{\svdots}
{\vbox{\baselineskip3\p@ \lineskiplimit\z@
       \hbox{.}\hbox{.}\hbox{.}}}
\makeatother

\title{Topology-Optimized Pneumatic Soft Actuator:\\ Design and Experimental Validation}

{}
\author[ ]{Sumit Mehta}
\author[*]{Konstantinos Poulios}

\affil[ ]{Department of Civil and Mechanical Engineering \protect\\
Technical University of Denmark, Denmark}

\date{}
\begin{document}
\maketitle

\vspace{-13mm}
\begin{abstract}
This paper demonstrates the computational design of soft elastomeric pneumatic actuators using nonlinear topology optimization.
An existing density- and porohyperelasticity-based topology optimization framework was extended from 2D to 3D and used to generate two manufacturable actuator designs, which were then studied numerically and experimentally.
For both designs, the objective was to maximize the bending response for a prescribed actuation pressure under two different allowable strain limits.
A key advantage of the employed topology optimization framework is that it can consistently, during the optimization, account for the very large deformations induced upon pressurization.
The two optimized 3D designs were fabricated using stereolithography and experimentally tested to validate their performance.
\end{abstract}

\let\thefootnote\relax\footnote{\hskip -0.5cm\scriptsize $^{\rm *}$Corresponding author: kopo@dtu.dk}

\section{Introduction}
In recent years, soft robotics has gained popularity owing to its inherent compliance, which resembles the mechanical characteristics of living organisms.
Unlike traditional rigid robots, soft robotic devices are typically fabricated from polymeric gels or elastomers.
These materials make them lightweight, cost-effective, and well-suited for navigating complex environments while ensuring safe interaction with sensitive components such as biological tissues.
Soft robotic actuators have found extensive applications in various domains, including soft grippers \citep{2011Il-Ma-Sh-Ch-Wh, 2018Sh-Ca-Fl-Sh}, locomotion \citep{2020Ch-Ca-Sa-Yu-Do-Ta-Ca}, human assistance \citep{2021Ar-St-Pe-Ci, 2015Po-Wa-Ga-Wo-Wa}, and minimally invasive surgery \citep{2019Ru-Da-My, 2018Hu-Ch-Lu-Zh-Zh}.
Among many possible actuation methods, soft robotic devices can be electrically driven \citep{2016Um-Vi-Tr, 2024Li-Li-Wu-Zh-Qu-Yu}, driven by magnetic fields \citep{2022Wa-Zh-Ji-We-Bi-Zh}, or actuated by means of pneumatic or hydraulic pressure \citep{2014Mo-Po-Ke-We-Sh-Gu-Sh-Be-Wa-Wh, 2015Po-Wa-Ov-Ga-Wo-Be-Wa}.
Pneumatic actuation is a particularly attractive option due to its low cost, simplicity, and rapid response \citep{2022Xa-Ta-Zo-Pi-HoYo-La-Ha-Yo-Bo}.

Soft robotic devices are traditionally developed based on designer intuition, extensive manual experimentation, and biological inspiration \citep{2013Ki-La-Tr,2015Po-Wa-Ov-Ga-Wo-Be-Wa,2024Li-Li-Wu-Zh-Qu-Yu,2025Sa-Ul-Is}, offering limited assurance in terms of overall functional performance.
This limitation applies to all actuation mechanisms in general, but also, at least until recently, to pneumatic actuators in particular \cite{2022Xa-Ta-Zo-Pi-HoYo-La-Ha-Yo-Bo}.
Recent developments, however, have enabled the design of optimized pneumatic actuators through computational methods instead.
Parametric optimization was used for example in~\cite{2020Gu-Ka}, to design pneumatically activated crawling soft robots.
Shape optimization was successfully used in~\cite{2023Ch-So-Ch-Gu-Zh} to design pneumatic actuators with optimized performance in extension, twisting, and bending.
Compared to parametric and shape optimization, topology optimization offers greater design freedom, yet it comes with an additional challenge.
During optimization, holes may appear in the design, leading to leakage of the pneumatic fluid.
To address this issue, the authors of \cite{2020So-Si} employed a three-valued density field with values -1, 0, and 1, representing pressurized fluid, solid, and non-pressurized fluid respectively, combined with specialized projection and filtering techniques.
Using this approach, they successfully generated 2D designs of pneumatic actuators for bending and extension, albeit limited by the assumption of small deformations.
An alternative approach relies on the use of a boundary identification algorithm in order to apply design-dependent pressure loads in a topology optimization framework \cite{2022Lu-To,2025Ch-Li}.
Both works applied this method alongside small-deformation topology optimization to obtain 2D designs for pneunet-type bending actuators.
Subsequently, the actuator responses under large deformations were evaluated through nonlinear post-simulation and experimental testing.
A different approach was proposed in \cite{2020Ku-Fr-La,2021Ku-La}, based on poroelasticity and a drainage term in solid regions, which was used to design pressure-actuated compliant mechanisms in 2D and 3D, also under the small-deformation assumption.
Some optimized pneunet-type 2D designs produced with this approach have recently been experimentally validated \cite{2026Ku-Pr-Pi-Ho-La}.
Common to all these works is that optimization was performed using small-deformation theory, while only the post-evaluation of the design was conducted with a large-deformation-capable finite element model.

One of the first works where large-deformation theory was actually used in the optimization of a pneumatic actuator can be found in \cite{2020Ca-Po-Ni}.
Nonetheless, this work relies on an algorithmic identification of pressurized cavities which is, in general, not smoothly differentiable, a considerable disadvantage in the context of optimization.
In \cite{2024Da-Wa-To}, simultaneous shape and topology optimization was used, while also accounting for large deformations, to produce pneumatically actuated gripper designs in 2D.
This is an interesting compromise.
On one side, the topology of the pressurized cavities is predefined, on the other side, their shape is still allowed to change.
A more general solution was proposed in \cite{2025Me-Po} by extending the small-deformation poroelasticity based topology optimization framework from \cite{2020Ku-Fr-La} to large deformations.
This was achieved mainly by employing the theory of porohyperelasticity from \cite{1996Si-Ka-Mc-Ba}.
The extended framework was used for producing 2D designs which were optimized while consistently accounting for the large deformations at the pressurized configuration.

The present work builds upon~\cite{2025Me-Po}, extending and applying the framework to 3D, and providing experimental validation.
As a benchmark problem, a cylindrical design domain was considered, with one end fixed and the other end rigidly connected to an actuated arm, whose motion is inhibited by a spring foundation.
This setup made it possible to generate pneumatic elastomer actuator designs optimized for bending response.
Optimizations were performed for given elastomer material properties, pneumatic pressure, foundation spring stiffness, and allowable strain limit.
Two optimized designs, obtained for different levels of allowable strain, were fabricated using silicone resin-based stereolithography (SLA) 3D printing.
The printed actuators were then tested experimentally and their performance was compared against the numerical predictions, in order to assess the overall efficacy of the proposed design methodology.

The remainder of the paper is structured as follows.
Section~2 discusses the numerical methodology and the optimization framework for designing the pneumatic soft actuator.
Section~3 presents the two optimization cases and the resulting actuator designs that were obtained for achieving maximum bending under pressurization.
Section~4 describes the experimental methodology, detailing the actuator fabrication method and the utilized test bench including its construction, sensor calibration procedures, and data acquisition techniques.
In Section~5, the numerical response of the actuator is compared with the experimental measurements, followed by a discussion about the efficacy of the proposed computational design approach.
Finally, concluding remarks are presented in Section~6.

\section{Numerical Methodology}
This section briefly describes the porohyperelasticity-based topology optimization framework for pneumatic soft actuators from~\citep{2025Me-Po}, and elaborates on its extension to 3D and other adaptations for the specific benchmark problem treated in the present work.

\subsection{Design parametrization}
At the core of the employed density-based topology optimization framework is a porohyperelastic ersatz material.
In void regions of some design domain $\Omega$, corresponding to cavities, the material has extremely high permeability and extremely low stiffness.
In solid regions, permeability is very low and the stiffness corresponds to the actual elastomer material.
The presence of void or solid at any point in the domain is controlled by a scalar level-set field $\chi$, which serves as the design variable to be optimized.
To link the level-set field $\chi$ with the ersatz material, the former is first mapped to a physical density field $\rho$, via the sigmoid function 
\begin{equation}
\rho{(\chi)} = \dfrac{1}{1 + e^{- \chi}},
\label{eq:level_set_fnc}
\end{equation}
which ensures that $\rho$ remains bound within the interval $(0,1)$, \citep{2021Bl-Si-Po}.
Then standard RAMP interpolation \citep{2001St-Sv} is applied on $\rho$ to obtain the bulk and shear moduli of the ersatz material as
\begin{equation}
\begin{Bmatrix}K(\chi)\\G(\chi)\end{Bmatrix}
=
\begin{Bmatrix}K_\mathrm{s}\\G_\mathrm{s}\end{Bmatrix}
\left(
\mathcal{E}_0 + (1 - \mathcal{E}_0) \dfrac{\rho(\chi)}{1 + \hat{p} (1 - \rho(\chi))}
\right),
\label{eq:ramp_interp}
\end{equation}
where $K_\mathrm{s}$ and $G_\mathrm{s}$ are the initial bulk and shear moduli for the solid.
The RAMP penalization parameter $\hat{p}\!=\!3$ and the void-solid stiffness contrast $\mathcal{E}_0\!=\!10^{-6}$ are used throughout the present work, simply adopted from \citep{2025Me-Po}.

The permeability of the ersatz material is assumed to vary according to the interpolation function proposed in~\citep{2025Me-Po}, i.e.
\begin{equation}
k(\chi) = k_\mathrm{v}+(k_\mathrm{s}-k_\mathrm{v})
                       \operatorname{\rho}(\chi\!+\!8)
=\dfrac{k_\mathrm{s}+k_\mathrm{v}e^{-\chi-8}}{1+e^{-\chi-8}},
\end{equation}
where $k_\mathrm{v}$ is the permeability in the void, while the permeability of the solid is $k_\mathrm{s}\!=\!10^{-6}k_\mathrm{v}$.

To avoid diffusion of pressurized fluid through solid walls, the formulation also uses a fictitious drainage term in the solid with intensity proportional to the density field, 
\begin{equation}
Q_{\mathrm{out}}\!(\chi)=Q_\mathrm{s}\,\rho(\chi).
\end{equation}
This term  is negligible in the void, while its value in the solid tends to the constant $Q_\mathrm{s}\!=\!\left(-\ln(0.1)/(0.02\,L_i)\right)^2k_\mathrm{s}$, according to~\citep{2025Me-Po}, where $L_i$ is the prescribed width of the solid-void interface, to be introduced later.

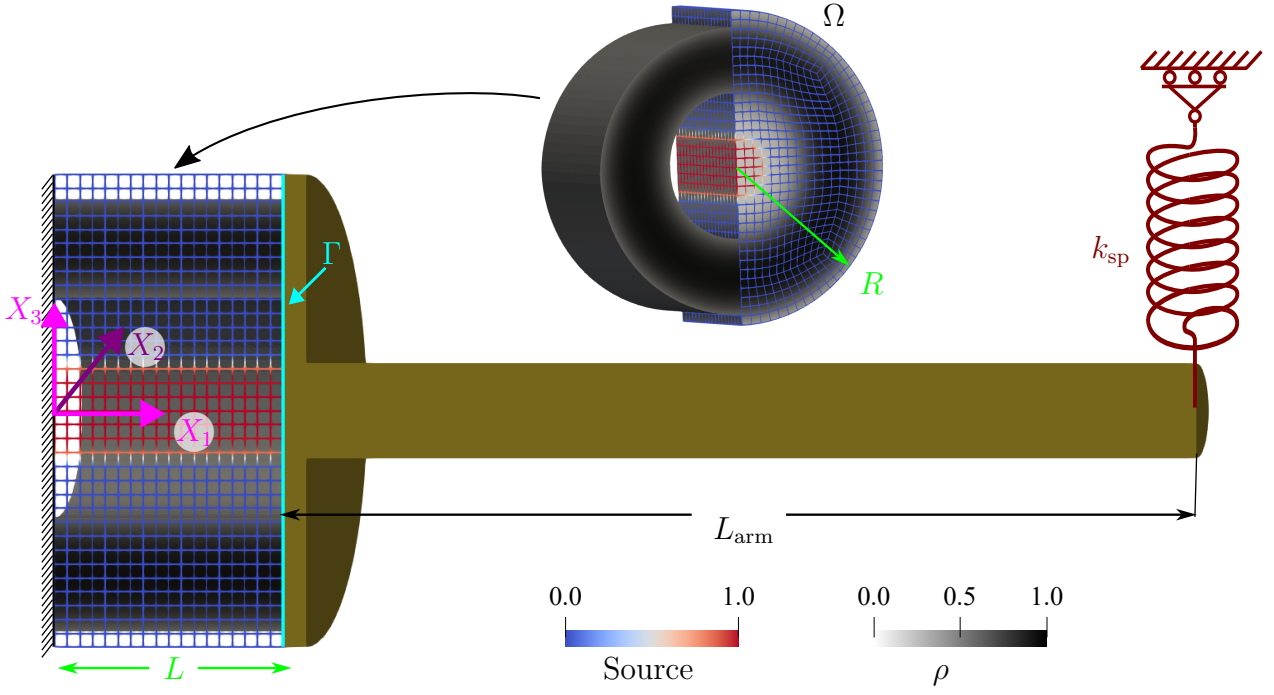
\begin{figure}[!h]
  \centering
  \begin{overpic}[width=0.95\linewidth]
                 {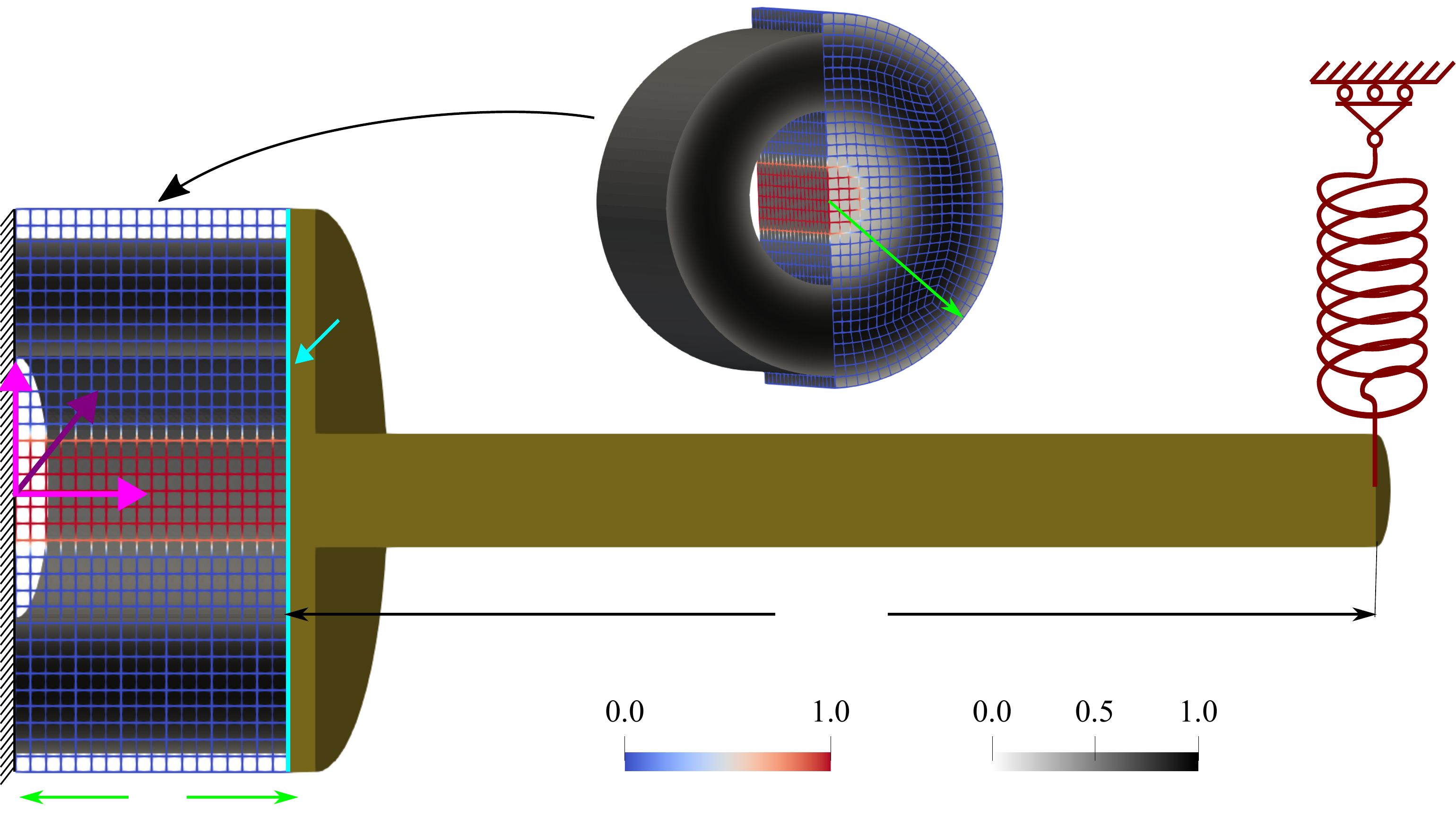}
  \put(10,0.5){$\color[rgb]{0,1,0} L$}
  \put(67,32){$\color[rgb]{0,1,0} R$}
  \put(55,12){$L_\mathrm{arm}$}
  \put(-3,30){$\color[rgb]{1,0,1} X_3$}
  \put(12.5,20.7){\transparent{0.7}\color{white}\circle*{12mm}}
  \put(11,20){$\color[rgb]{1,0,1} X_1$}
  \put(8.5,27.7){\transparent{0.7}\color{white}\circle*{20mm}}
  \put(7,27){$\color[rgb]{0.5,0,0.5} X_2$}
  \put(46,0.5){Source}
  \put(73,0.5){$\rho$}
  \put(86,35){$\color[rgb]{0.5,0,0} k_\mathrm{sp}$}
  \put(64,54){$\Omega$}
  \put(23,34.5){$\color[rgb]{0,1,1}\Gamma$}
  \end{overpic}
  \caption{Design domain $\Omega$ with illustrations of the initial design $\rho(\chi_0)$ thresholded at $\rho\!=\!0.5$, the pressurized region (source), the actuated rigid arm, and the foundation spring.}
  \label{fig:init_design}
\end{figure}

\subsection{Governing physics}
Let us now consider the cylindrical domain $\Omega$ from Figure~\ref{fig:init_design}, with radius $R$ and length $L$, cut in half, to exploit symmetry.
The design field $\chi$ is defined over $\Omega$, and is for now considered as given.
Moreover, a displacement field $\uu$ and a pressure field $p$ are also defined in $\Omega$, according to the theory of porohyperelasticity from \citep{1996Si-Ka-Mc-Ba}.
Combining this theory with the Neo-Hookean constitutive law from \cite{1985Si-Ta-Pi}, provides the necessary equations that govern the mechanical behavior of the system, including the propagation of the pressurized fluid through all connected cavities.
The equilibrium equation governing the pressure field $p$ is
\begin{equation}\label{eq:Darcy_flow}
\begin{aligned}
\RR_p(\chi,p,\uu)\!\left[\delta p\right]=
\int_{\Omega}
 &~~k(\chi) \detF
    \left(\F^{-1}\F^{-T}\nabla p \right)
    \!\cdot\! \nabla\delta p
\\
 &+\big(Q_{\mathrm{in}}(\XX)\left(p\!-\!p_{\mathrm{in}}\right)
        +Q_{\mathrm{out}}\!(\chi)\left(p\!-\!p_{\mathrm{out}}\right)
   \big)\delta p
~d\Omega= 0~~~\forall~\delta p,
\end{aligned}
\end{equation}
where $\F\!=\!\II\!+\!\nabla\uu$ is the deformation gradient.
The reference ambient pressure~$p_\mathrm{out}$ is actually set to zero.
The fluid source function $Q_{\mathrm{in}}(\XX)$ is nonzero only at locations $\XX$ within the core region of the domain $\Omega$, considered as the inlet region, where the actuation pressure~$p_\mathrm{in}$ is provided.
The exact definition of $Q_{\mathrm{in}}(\XX)$ can be found in Appendix~\ref{appendix_Qin}.

The mechanical equilibrium equation governing the displacement field $\uu$ is
\begin{equation}\label{eq:mech_equilibrium}
\begin{aligned}
\RR_\uu(\chi,p,\uu,\qq)[\ddelta\uu]=\!\!
\int_{\Omega}\!
  \left(\left(K(\chi) \ln\!\detF\!-\!\detF p\right) \II
              \!+\!G(\chi) \detF^{-2/3}\Dev{\F\F^T}\right)
    \!:\!\left(\nabla\ddelta\uu\,\F^{-1}\right)&
    \\[0pt]
    +~c_r ~\mathbb{H} \uu  \svdots \mathbb{H} \ddelta\uu
    ~&d\Omega
\\
+\int_{\Gamma} \qq \cdot \ddelta\uu~d\Gamma
= 0
~~~~~~~~~~~~~~~~~~~~~~~~~~~~~~~~~~~~~~~~~~~~~~
~~~~~~~~~~~~
&~\forall~\ddelta\uu,
\end{aligned}
\end{equation}
where $c_r$ is a regularization parameter for penalizing the Hessian of the displacement field in the interior of finite elements.
This is to avoid excessive element distortion close to the pressure source.
The traction field $\qq$ is defined on the side $\Gamma$ of the domain $\Omega$, which is attached to the rigid arm.
It serves as a Lagrange multiplier field that connects the porohyperelastic domain with the rigid arm.

On its opposite end, the rigid arm is attached to a vertical linear spring with stiffness $k_\mathrm{sp}$.
Due to symmetry, the arm is neither supposed to be displaced in the $X_2$ direction, nor to rotate about the $X_1$ or $X_3$ axes.
The remaining two translations and one rotation are included in a vector $\TT\!=\!\left[T_1,T_3,\theta_2\right]$, which describes the placement of the rigid arm, with origin at $\XX\!=\!\left[L,0,0\right]$.
The equilibrium equations for the vector field $\qq$ as well as for the vector $\TT$, including the force from the linear spring, are
\begin{equation}
\RR_\qq(\uu,\TT)[\ddelta\qq]=
\int_\Gamma
\ddelta\qq \cdot
\left(\uu-\begin{pmatrix}T_1+X_3\sin\theta_2\\
                         0\\
                         T_3-X_3(1-\cos\theta_2)
          \end{pmatrix}\right)
d\Gamma=0~~~\forall~\ddelta\qq,
\end{equation}
\begin{equation}
\begin{aligned}
\RR_\TT(\qq,\TT)[\ddelta\TT]=
k_\mathrm{sp}\!
\left(T_3\!-\!L_\mathrm{arm}\sin\theta_2\right)
&
\left(\delta T_3\!-\!L_\mathrm{arm}\cos\theta_2\,\delta\theta_2\right)\\
-&
\int_\Gamma
\qq\cdot
\begin{pmatrix}\delta T_1+X_3\cos\theta_2\,\delta\theta_2\\
               0\\
               \delta T_3-X_3\sin\theta_2\,\delta\theta_2
\end{pmatrix}
d\Gamma=0~~~\forall~\delta T_1,\delta T_3,\delta\theta_2.
\end{aligned}
\end{equation}

\subsection{Optimization equations}
So far, the level-set $\chi$, that defines the geometry of the actuator, was considered as given.
The computational design process consists in finding such a field $\chi$ that represents a manufacturable structure and maximizes some performance metric of interest.
In our benchmark from Figure~\ref{fig:init_design}, the objective is to maximize the elongation of the foundation spring for a given actuation pressure $p_\mathrm{in}$.
In particular, we aim at minimizing the objective function
\begin{equation}
C_0(\TT)=\dfrac{T_3}{L_\mathrm{arm}}-\sin\theta_2+\dfrac{1000}{2}\left<\dfrac{T_3}{L_\mathrm{arm}}\right>^2, \label{eq:main_obj}
\end{equation}
which depends only on components of the rigid arm displacement vector $\TT$.
The last term, which penalizes the positive part of $T_3$, is just for ensuring enough transverse stiffness of the structure by not letting the origin of the rigid arm move upwards.
The remaining part of $C_0$, involving $T_3$ and $\theta_2$, corresponds to the vertical displacement of the spring end.
Minimizing the vertical position of the displaced spring end corresponds to a stronger bending response.

According to~\cite{2025Me-Po}, additional terms need to be included in the objective in order to ensure manufacturability of the optimized design described by $\chi$, to have consistent gradient information, and to avoid certain pitfalls during the optimization.
In particular, each optimized actuator of the present work corresponds to a stationary point of the augmented objective function
\begin{equation}\label{eq:aug_objfunc}
\begin{aligned}
C^*=&~C_0(\TT)+C_\chi(\chi)+C_A(\chi)+C_i(\chi)+C_p(p)+C_\Psi(\chi,\uu)+C_t(\dot{\chi})\\
   &+\RR(\chi,p,\uu,\qq,\TT)\left[\Lambda_p,\LLambda_\uu,\LLambda_\qq,\LLambda_\TT\right],
\end{aligned}
\end{equation}
where $\Lambda_p$, $\LLambda_\uu$, $\LLambda_\qq$, and $\LLambda_\TT$ are adjoint variables, respective to the previously introduced variables $p$, $\uu$, $\qq$, and $\TT$.
These adjoint variables are necessary for obtaining a consistent gradient of the objective function with respect to the design field $\chi$.
The last term in Eq.~\eqref{eq:aug_objfunc} is simply the sum of all previously defined residual equations, i.e. $\RR=\RR_p\!+\!\RR_\uu+\!\RR_\qq+\!\RR_\TT$.

The term $C_\chi$ in $C^*$ simply penalizes any solid material in any region that is prescribed as passive void region.
Its exact definition can be found in Appendix~\ref{appendix_Cx}.
The term $C_A$ penalizes the total surface area of the design, in order to promote simpler designs as discussed in~\citep{2025Me-Po}.
The term~$C_i$ limits the slope of the field $\chi$ and thereby imposes a minimum width $L_i$ for the solid-to-void transition zone.
The term~$C_p$ penalizes the occurrence of pressure leaks during the optimization design updates.
The term~$C_\Psi$ penalizes designs that violate a prescribed allowable strain energy density limit, with the main purpose of avoiding designs that involve ballooning~\citep{2023Ga-Ku-Si}.
Finally, the term $C_t$ which depends on the rate $\dot{\chi}$ of the design field, enables performing the optimization as a sequence of evolving design updates.
The optimization problem formulation is completed by defining all remaining objective function contributions
\begin{equation}\label{eq:C_A}
C_A(\chi)= \dfrac{c_A}{\Det{\Omega}^{2/3}}
\dfrac{8}{L_i}\int_\Omega \rho(\chi)\left(1-\rho(\chi)\right) d\Omega,
\end{equation}

\begin{equation}\label{eq:C_i}
C_i(\chi) =
\int_{\Omega}
  \dfrac{c_i}{6}
  \Bigl\langle\Norm{\nabla\chi}-8/L_i\Bigr\rangle^{\!6}
d\Omega,
\end{equation}

\begin{equation}\label{eq:C_p}
C_p(p)=\dfrac{1}{\Det{\Gamma_\mathrm{ext}}}
       \int_{\Gamma_\mathrm{ext}}
       \dfrac{c_p}{2}\left(\dfrac{p-p_\mathrm{out}}
                                 {p_\mathrm{in}-p_\mathrm{out}}
                     \right)^2
       d\Gamma,
\end{equation}

\begin{equation}\label{eq:C_Psi}
C_\Psi(\chi, \uu) =
\int_{\Omega}
  \dfrac{c_\Psi}{6}
  \Bigl\langle
    \dfrac{\Psi\!\left(\chi\!+\!4/L_i,\uu\right)}
          {\Psi_\mathrm{lim}}
    -1
  \Bigr\rangle^{\!6}
d\Omega,
\end{equation}
and with $C_t(\dot{\chi})$ approximated for some pseudo-time step $\Delta t$ as
\begin{equation}\label{eq:design_damping}
C_t(\chi,\chi_\mathrm{old}) = \int_{\Omega}
      \dfrac{\left(\chi-\chi_\mathrm{old}\right)^2
             +L_t^2 \Norm{\nabla\chi-\nabla\chi_\mathrm{old}}^2}
            {2\Delta t^2}
      ~d\Omega.
\end{equation}
The constants $c_A$, $c_i$, $c_p$, and $c_\Psi$ control the relative weighing for the respective terms in the objective function.
The solid-void interface width parameter $L_i$ introduces a length-scale by limiting the level-set steepness through the penalty term from Eq.~\eqref{eq:C_i}.
The leaking pressure constraint in Eq.~\eqref{eq:C_p} is applied to the external cylindrical surface $\Gamma_\mathrm{ext}$ of the domain $\Omega$, with inlet and ambient pressure parameters $p_\mathrm{in}$ and $p_\mathrm{out}$ already previously defined in Eq.~\eqref{eq:Darcy_flow}.
The strain energy density function $\Psi$ in Eq.~\eqref{eq:C_Psi} is for the hyperelastic law from~\citep{1985Si-Ta-Pi} also used in Eq.~\eqref{eq:mech_equilibrium}, and $\Psi_\mathrm{lim}$ is the allowable strain energy density limit.
Finally, in Eq.~\eqref{eq:design_damping}, the update from the last known design $\chi_\mathrm{old}$ to the next design $\chi$ corresponds to a step $\Delta t$ in design pseudo-time, and $L_t$ is a length scale that controls the smoothness of the design updates.

Having defined the augmented objective function $C^*$ from Eq.~\eqref{eq:aug_objfunc}, it is straightforward to obtain the optimality condition in the form of the weak system of equations
\begin{align}\label{eq:optimality_adjoint_eqs}
  C_{\chi,\chi}\!\left[ \delta\chi \right]
 +C_{A,\chi}   \!\left[ \delta\chi \right]
 +C_{i,\chi}   \!\left[ \delta\chi \right]
 +C_{\Psi,\chi}\!\left[ \delta\chi \right]
 +C_{t,\chi}   \!\left[ \delta\chi \right]
&+\RR_{,\chi}  \!\left[\Lambda_p,\LLambda_\uu{\color{gray!40},\LLambda_\qq,\LLambda_\TT}\right]
                 \left[ \delta\chi \right]
 \nonumber\\
 +~C_{p,p}     \!\left[ \dLambda_p \right]
&+\RR_{,p}     \!\left[ \Lambda_p,\LLambda_\uu{\color{gray!40},\LLambda_\qq,\LLambda_\TT}\right]
                 \left[ \dLambda_p \right]
 \nonumber\\
 +~C_{\Psi,\uu}\!\left[ \ddLambda_\uu \right]
&+\RR_{,\uu}   \!\left[\Lambda_p,\LLambda_\uu,\LLambda_\qq{\color{gray!40},\LLambda_\TT}\right]
                 \left[ \ddLambda_\uu \right]
 \nonumber\\
&+\RR_{,\qq}   \!\left[ {\color{gray!40}\Lambda_p,}\,\LLambda_\uu,
                       {\color{gray!40}\LLambda_\qq,}\,\LLambda_\TT\right]
                 \left[ \ddLambda_\qq \right]
 \nonumber\\
 +~C_{0,\TT}   \!\left[ \ddLambda_\TT \right]
&+\RR_{,\TT}   \!\left[ {\color{gray!40}\Lambda_p,\LLambda_\uu,}\,\LLambda_\qq,\LLambda_\TT\right]
                 \left[ \ddLambda_\TT \right]
               = 0.
\end{align}
This condition has to be satisfied for any $\delta\chi$, $\dLambda_p$, $\ddLambda_\uu$, $\ddLambda_\qq$, $\ddLambda_\TT$, at any design update $\chi_\mathrm{old}\rightarrow\chi$ of size $\Delta t$.
The adjoint variables, $\Lambda_p$, $\LLambda_\uu$, $\LLambda_\qq$, and $\LLambda_\TT$ as well as their variations in Eq.~\eqref{eq:optimality_adjoint_eqs}, are assumed to be of the same sizes as the respective physics variables $p$, $\uu$, $\qq$, and~$\TT$.

The first row of Eq.~\eqref{eq:optimality_adjoint_eqs} corresponds to the actual optimality condition, while the remaining four rows correspond to the adjoint equations that are solved for $\Lambda_p$, $\LLambda_\uu$, $\LLambda_\qq$, and $\LLambda_\TT$.
These four equations are per definition linear with respect to the adjoint variables, but they include nonlinear dependencies on the design variable and the primary physics variables.
Non-applicable dependencies on adjoint variables, in the derivatives of the residual $\RR$, are noted in Eq.~\eqref{eq:optimality_adjoint_eqs} with gray font color.

Further details about the formulation and the solution process are provided in \cite{2025Me-Po}.
Here, only specific choices regarding this 3D adaptation of the method will be provided.
The 3D computational domain, with the form of a half cylinder, was discretized with \nth{1} order 8-node hexahedral elements in the mesh shown in Figure~\ref{fig:init_design}.
The level-set field $\chi$ was approximated with isoparametric finite elements, i.e. with 8-node hexahedral basis functions.
However, the fields~$\uu$ and~$\qq$ were discretized with \nth{2} order 20-node hexahedral element basis functions, i.e. subparametric elements.
For the $\uu$ field, all degrees of freedom at the fixed side were eliminated, as well as the degrees of freedom on the symmetry plane that have direction parallel to the $X_2$ axis.
For the $\qq$ field, only degrees of freedom at the rightmost side of the domain, i.e. at $X_1\!=\!L$, were kept, except those that lie on the symmetry plane and are parallel to the $X_2$ direction.
All integrals for the assembly of residuals and tangent matrices were computed with 14 integration points per hexahedral, according to the \nth{5} degree cubature rule from \cite{1971Str,2003Co}.
%https://nines.cs.kuleuven.be/ecf/mtables.html

\section{Numerical Results}
The model described so far was run with the data reported in Table~\ref{tab:model_params}, and with the initial design shown in Figure~\ref{fig:init_design}.
The initial design resembles a simple tube but has a small asymmetry in the $X_3$ direction, so that its upper side is slightly thinner than the lower side, in order to produce some small initial deflection of the actuated arm.
The exact mathematical definition of the initial design is of secondary importance, and it is simply provided in Appendix~\ref{appendix_init_design}, for the sake of completeness.

\begin{table}[b!]
\centering
\caption{Default model and discretization parameters for optimized pneumatic soft actuator.}\vspace{3mm}
\begin{tabular}{lll}
\hline\\[-12pt]
  $L\!=\!R\!=\!15$
        & Domain length and radius
        & [\si{\milli\meter}] \\
  $L_\mathrm{arm}\!=\!4\,R$
        & Length of actuated rigid arm
        & [\si{\milli\meter}] \\
  $E\!=\!2.736,~\nu\!=\!0.48$
        & Initial elasticity parameters for the solid
        & [\si{\newton\per\milli\meter\squared},~\si{-}] \\
  $k_\mathrm{sp}\!=\!1.6\!\cdot\!10^{-2} E\,R^3/L_\mathrm{arm}^2$
        & Spring stiffness (for half of the actuator)
        & [\si{\newton\per\milli\meter}] \\
  $p_\mathrm{in}\!=\!2\!\cdot\!10^{-2}E$
        & Source pressure
        & [\si{\newton\per\milli\meter\squared}] \\[2pt]
\hline&\\[-12pt]
  $N_1\!\times\!N_{23}\!=\!18\!\times\!528$
        & Number of elements in $X_1$-axis and $X_2 X_3$-plane
        & [\si{-}] \\[2pt]
\hline&\\[-12pt]
  $c_A\!=\!10^{-3}$
        & Penalization of estimated total surface area
        & \si{-} \\
  $c_i\!=\!1$
        & Penalization of interface width constraint
        & [\si{-}] \\
  $c_p\!=\!10^6$
        & Penalization of leaking pressure constraint
        & [\si{-}] \\
  $c_\Psi\!=\!10^{6}$
        & Penalization of strain energy density constraint
        & [\si{-}] \\[2pt]
\hline&\\[-12pt]
  $L_i\!=\!3$
        & Void-solid interface width
        & [\si{\milli\meter}] \\
  $L_t\!=\!5 L_i$
        & Design diffusivity length
        & [\si{\milli\meter}] \\[2pt]
\hline&\\[-12pt]
  $k_\mathrm{v}\!=\!1000\left[\si[per-mode=reciprocal]{\per\second}\right]L^2/E$
        & Void permeability
        & [\si{\milli\meter^4\per\newton\per\second}] \\
  $k_\mathrm{s}\!=\!10^{-6}~k_\mathrm{v}$
        & Solid permeability
        & [\si{\milli\meter^4\per\newton\per\second}] \\[2pt]
\hline&\\[-12pt]
  $\Psi_\mathrm{lim}\!=\!\tfrac{1}{2}\,0.14^2E$
        & Max. allowable strain energy density
        & [\si{\newton\per\milli\meter\squared}] \\[3pt]
\hline
\end{tabular}
\label{tab:model_params}
\end{table}

\begin{figure}[b!]
  \centering
  \begin{overpic}[height=0.37\linewidth,trim=18mm 0mm 17mm 0mm,clip]
                 {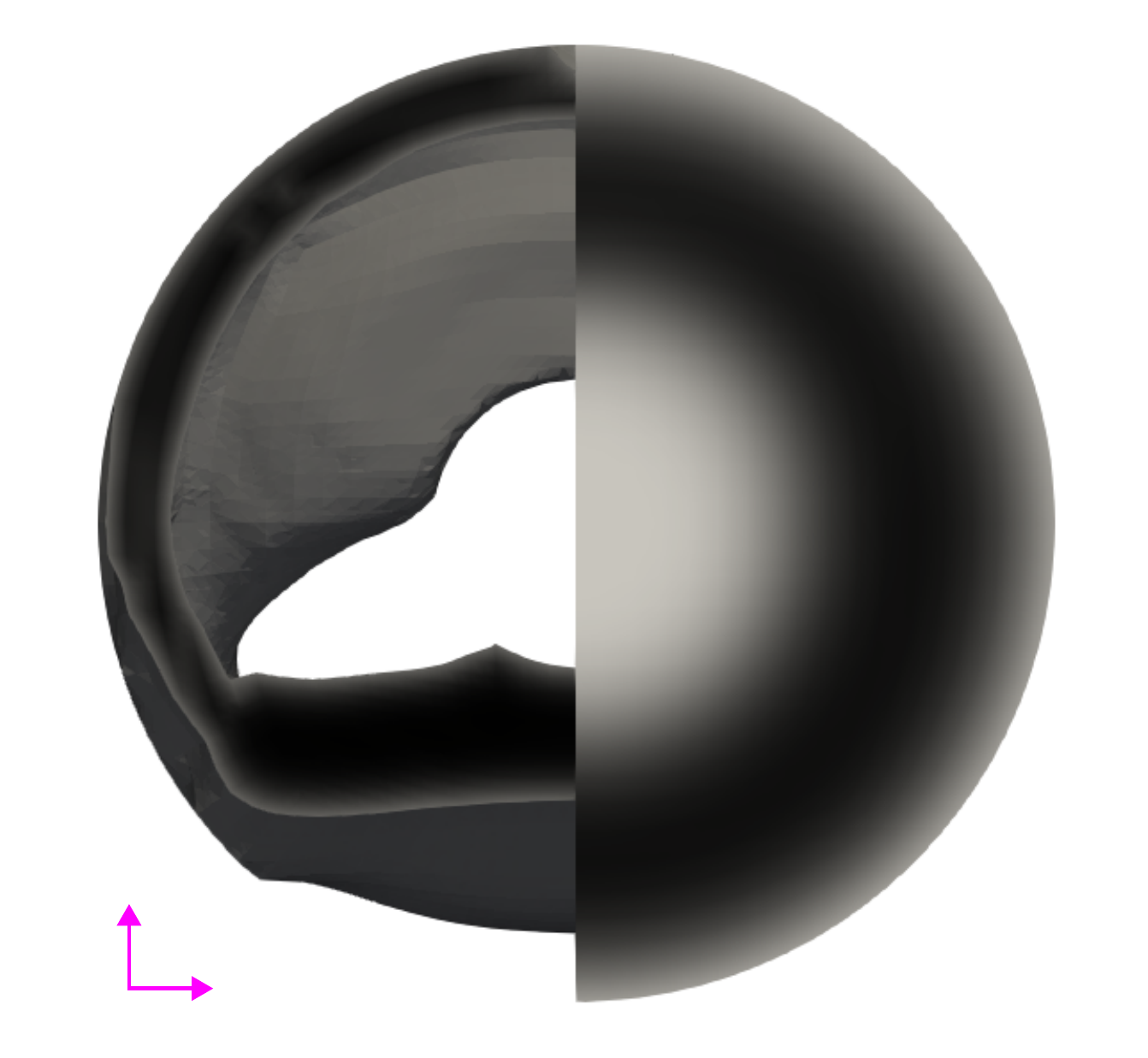}
    \put(0,95){a)}
    \put(46,50){\color{green}\vector(2,-1){40}}
    \put(71,41){\textcolor{green}{$R$}}
    \put(62,0){\footnotesize initial design}
    \put(78,6){\vector(-2,3){5}}
    \put(12,3){\color[rgb]{1,0,1}{$X_2$}}
    \put(1,15){\color[rgb]{1,0,1}{$X_3$}}
  \end{overpic}
  ~~~
  \begin{overpic}[height=0.37\linewidth,trim=0mm 0mm 0mm 0mm,clip]
                 {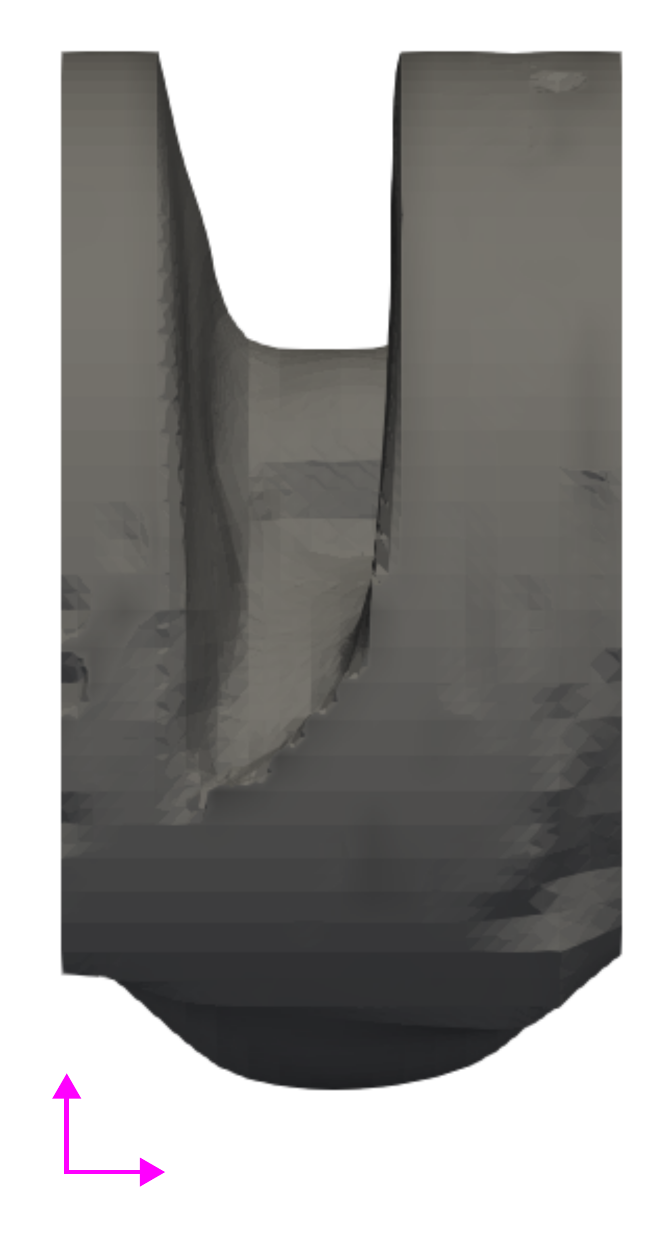}
    \put(-5,95){b)}
    \put(22,98){\color{green}\vector(-1,0){17}}
    \put(33,98){\color{green}\vector(1,0){17}}
    \put(25,96){\textcolor{green}{$L$}}
    \put(13,3){\color[rgb]{1,0,1}{$X_1$}}
    \put(0,15){\color[rgb]{1,0,1}{$X_3$}}
  \end{overpic}
  \begin{overpic}[height=0.37\linewidth,trim=0mm 0mm 0mm 0mm,clip]
                 {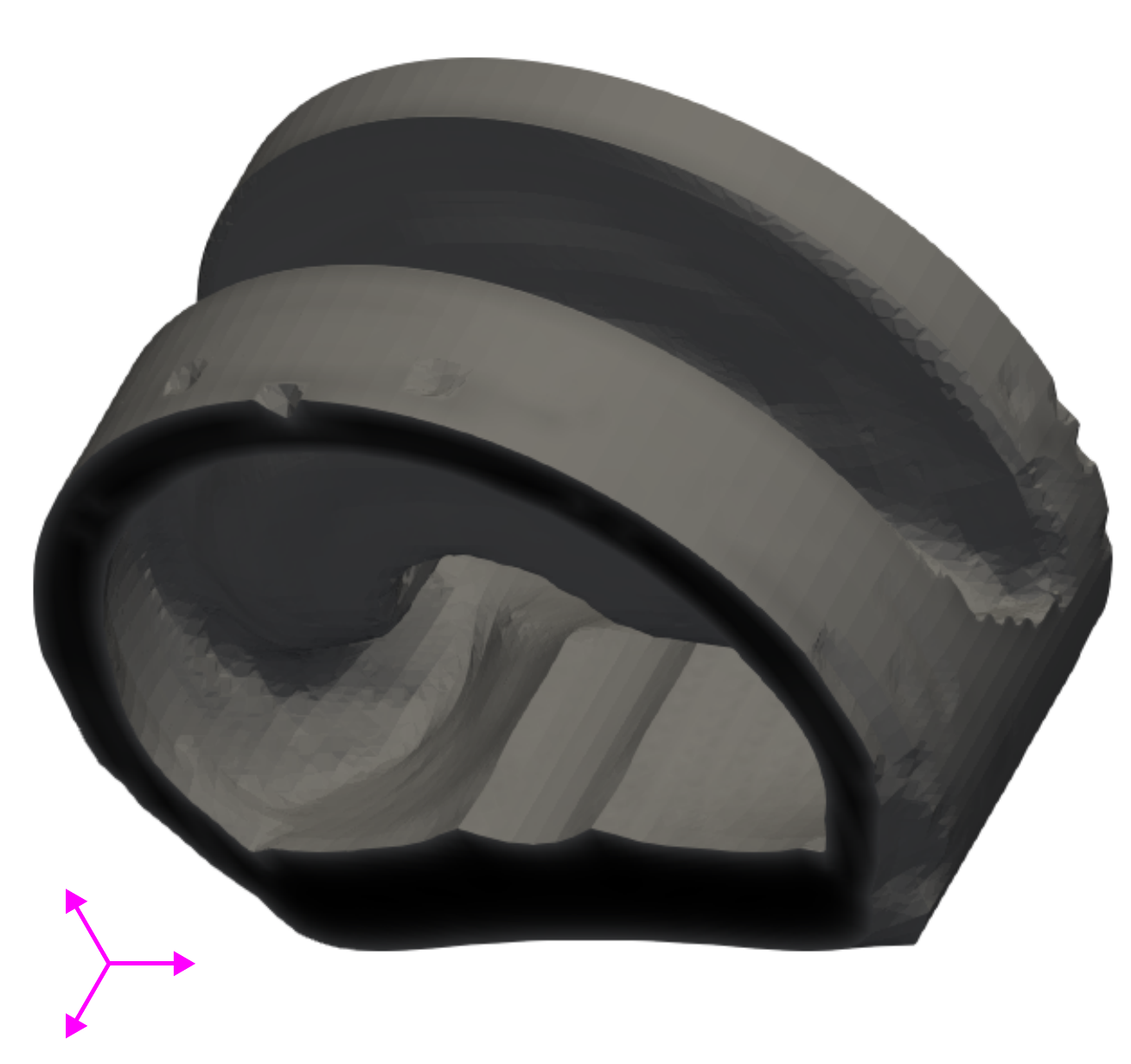}
    \put(3,87.5){c)}
    \put(-2,0){\color[rgb]{1,0,1}{$X_1$}}
    \put(15,2){\color[rgb]{1,0,1}{$X_2$}}
    \put(0,17){\color[rgb]{1,0,1}{$X_3$}}
  \end{overpic}
  \caption{Optimized 3D design obtained for $\sqrt{2 \Psi_\mathrm{lim}/E}\!=\!14\%$. a)~$X_1^-$~view of initial and final design, b) $X_2^+$~view, and c) isometric view of final design thresholded at $\rho\!=\!0.5$.}
  \label{fig:case14pct_3Dviews}
\end{figure}

\begin{figure}[t!]
\centering
\begin{overpic}[width=0.95\linewidth,trim=0cm 0cm 0cm 0cm,clip]
               {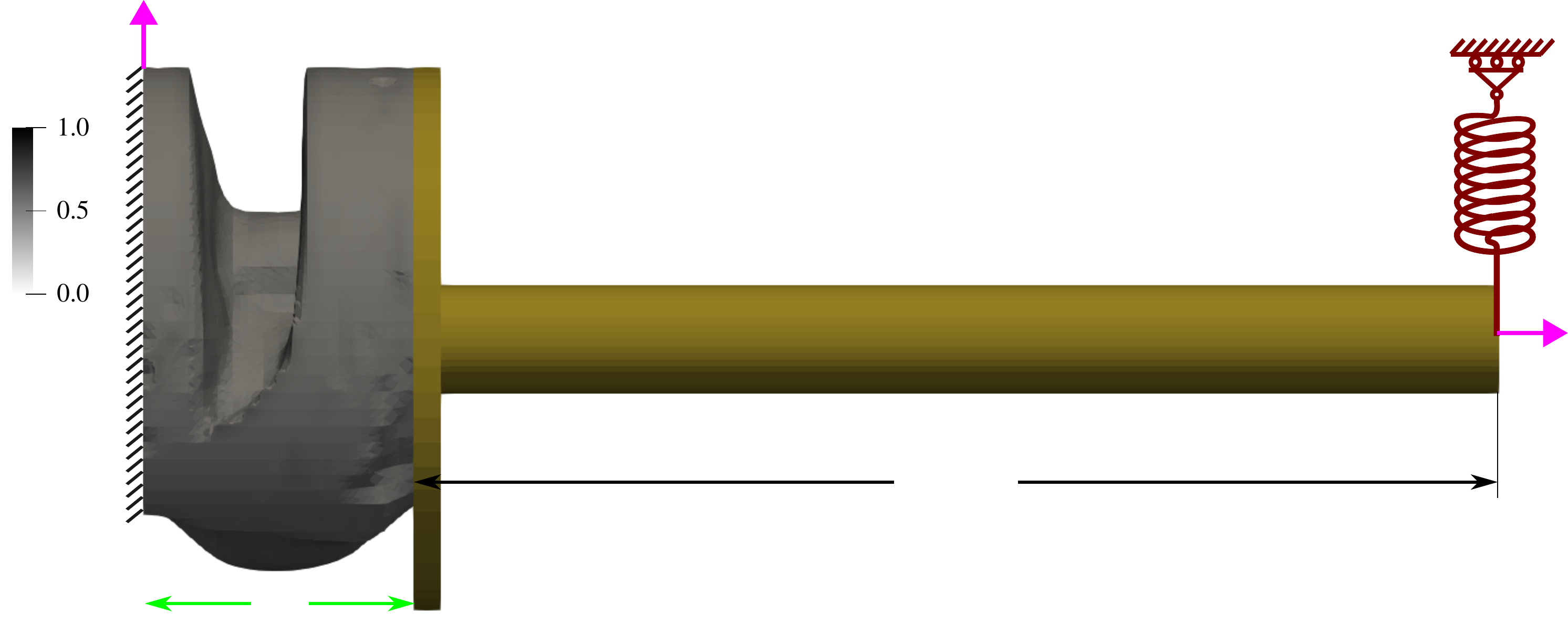}
  \put(0,35){$\rho(\chi)$}
  \put(17,1){$\color[rgb]{0,1,0} L$}
  \put(58,8.7){$L_\mathrm{arm}$}
  \put(98,15.5){$\color[rgb]{1,0,1} X_1$}
  \put(10,38.5){$\color[rgb]{1,0,1} X_3$}
\end{overpic}
\caption{Side view of the final design obtained for $\sqrt{2 \Psi_\mathrm{lim}/E}\!=\!14\%$, including support and actuated arm.}
\label{fig:case14pct_with_arm}
\end{figure}

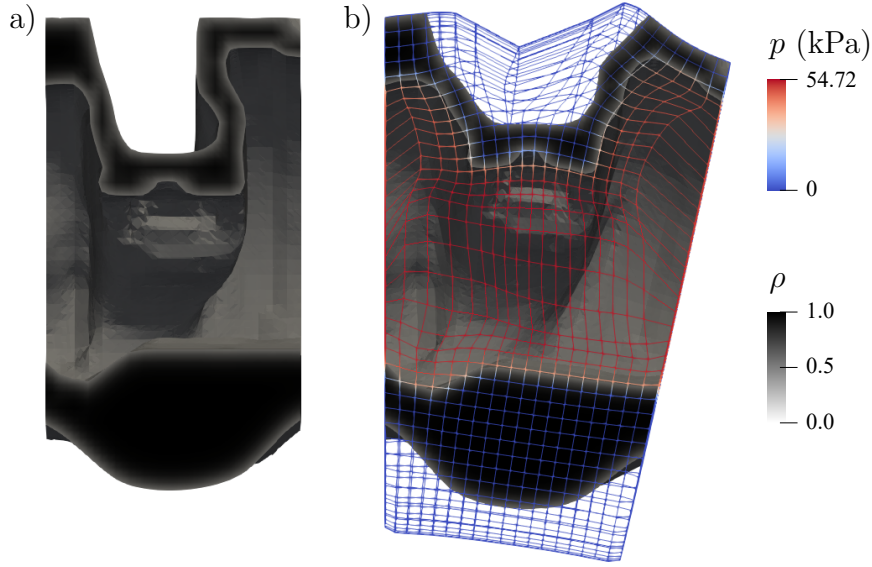
\begin{figure}[h!]
\centering
\begin{overpic}[width=0.65\linewidth,trim=0cm 0cm 0cm 0cm]
               {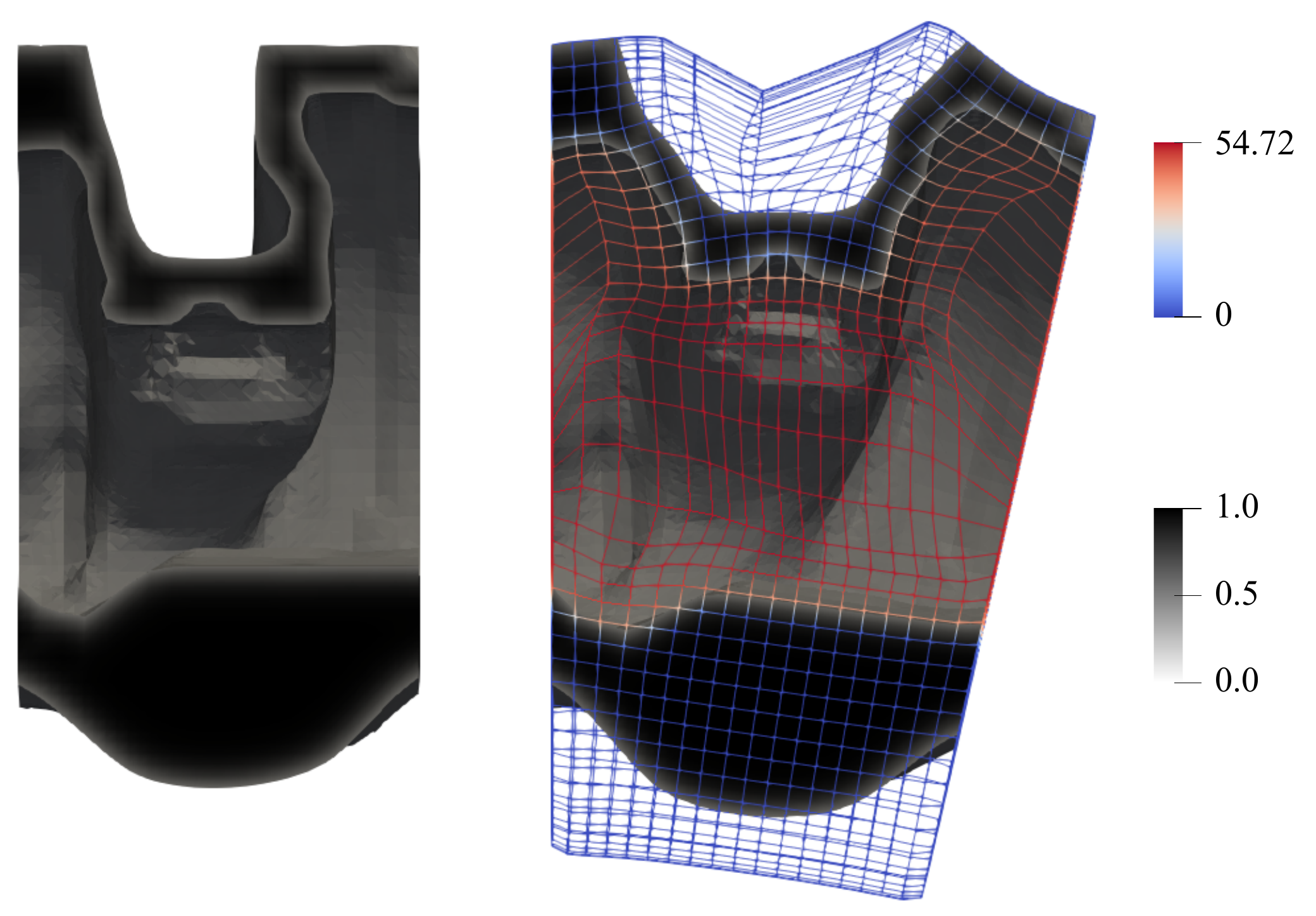}
  \put(88,62){$p ~ (\mathrm{kPa})$}
  \put(88,34){$\rho$}
  \put(-3, 65){a)}
  \put(37, 65){b)}
\end{overpic}
\caption{Section of final design obtained for $\sqrt{2 \Psi_\mathrm{lim}/E}\!=\!14\%$ in the undeformed (a) and deformed (b) configurations.}
\label{fig:case14pct_undef_def}
\end{figure}

Figure~\ref{fig:case14pct_3Dviews} shows the fully converged design produced with the data from Table~\ref{tab:model_params}.
In this visual, the second half of the symmetric design was reconstructed by mirroring, to better illustrate the actual component.
The grayscale colormap represents the density field $\rho$, with black corresponding to $\rho\!=\!1$ and the gray color on the surface corresponding to $\rho\!=\!0.5$.
The chosen strain energy density limit $\Psi_\mathrm{lim}$ would correspond to a strain of 14\% if the material was linear.
This rather strict limit for an elastomer was chosen because of the relatively low source pressure at 2\% of the elastomer's initial Young's modulus.
The reader is referred to \cite{2025Me-Po} for a parametric study with respect to source pressure and allowable strain energy density in~2D.

Figure~\ref{fig:case14pct_with_arm} shows a side view of the same actuator design, including the boundary condition and the actuated arm, in order to give a better impression of the optimized system.
Further details about this design are provided in Figure~\ref{fig:case14pct_undef_def}, which shows the interior of the actuator in its undeformed and deformed state.
The deformed state visualization includes also the deformed mesh, colored according to the pressure field $p$.
It is clear that a considerable rotation of the rigid arm can be produced for the prescribed actuation pressure at 2\% of the elastomer initial Young's modulus $E$, against an environment resistance at 1.6\% of the reference output stiffness $ER^3/L_\mathrm{arm}^2$.
In particular, the root point of the rigid arm is displaced downwards with $T_3\!=\!\SI{-2.305}{\milli\meter}\!=\!-0.1537 R$ and the arm is rotated about the $X_2$ axis with $\theta_2\!=\!\SI{0.21878}{\radian}\!=\!\SI{12.535}{\degree}$.

\begin{figure}[h!]
\centering
\begin{overpic}[width=0.516\linewidth,trim=0mm 4mm 0mm 0mm,clip]
               {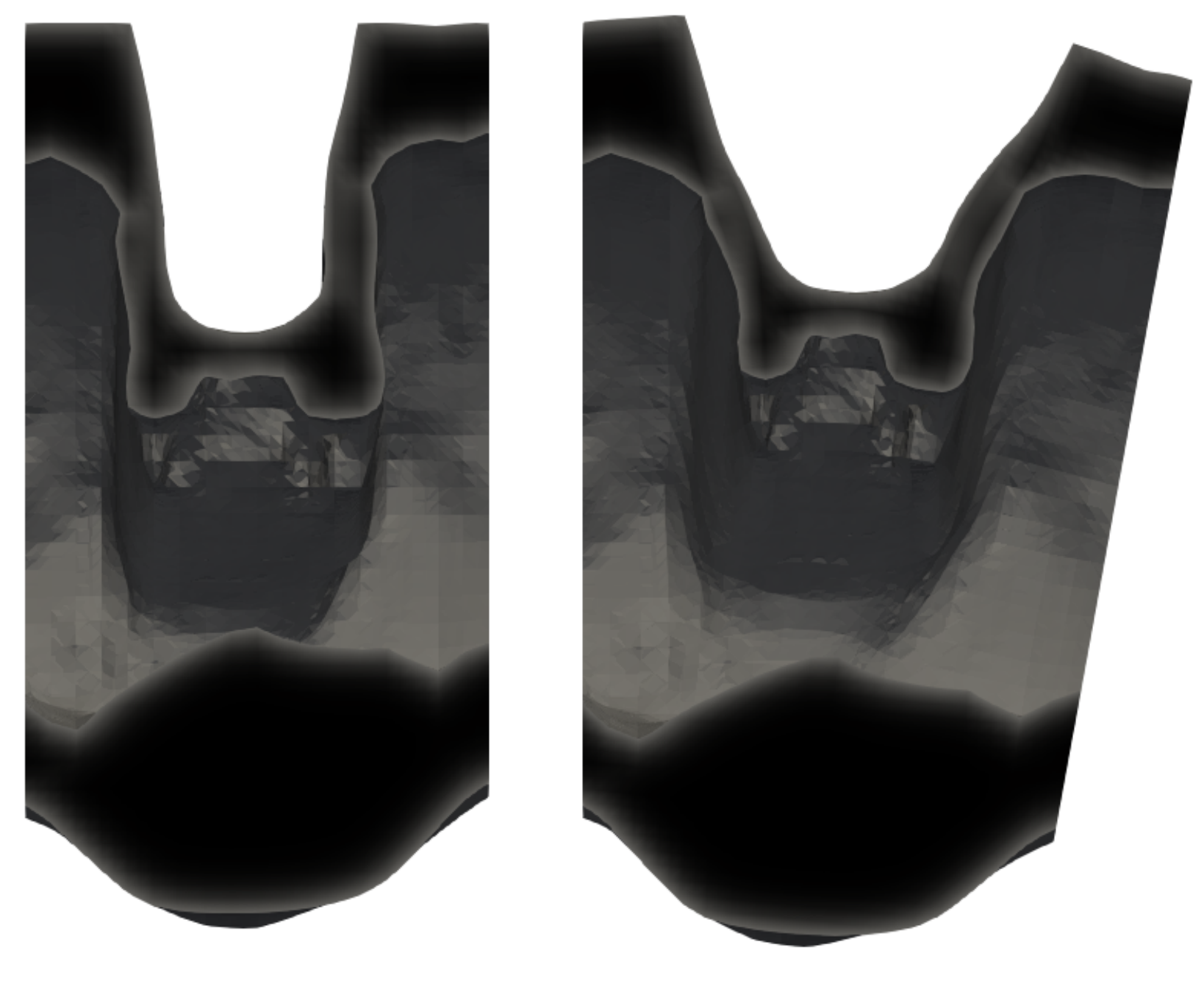}
  \put(-3, 76.5){a)}
  \put(43, 76.5){b)}
\end{overpic}
\caption{Section of final design obtained for $\sqrt{2 \Psi_\mathrm{lim}/E}\!=\!11\%$ in the undeformed (a) and deformed (b) configurations.}
\label{fig:case11pct_undef_def}
\end{figure}

\begin{table}[h!]
\centering
\caption{Performance of the two optimized soft actuators.}\vspace{3mm}
\begin{tabular}{cccc}
\hline\\[-12pt]
$\sqrt{2\Psi_\mathrm{lim}/E}$ & Arm rotation & Spring elongation & $C_0$\\[2pt]
\hline\\[-12pt]
14\% & \SI{12.535}{\degree}
     & $\SI{15.327}{\milli\meter}=1.0218R$
     & -0.255447\\
11\% & \SI{10.322}{\degree}
     & $\SI{12.376}{\milli\meter}=0.8251R$
     & -0.206272\\[2pt]
\hline
\end{tabular}
\label{tab:design_perfomance_comparison}
\end{table}

A second actuator design was produced with a stricter deformation constraint, i.e. with a lower allowable strain energy density limit $\Psi_\mathrm{lim}\!=\!\tfrac{1}{2}0.11^2E$, corresponding to 11\% strain of a linear equivalent material.
The fully converged design is shown in Figure~\ref{fig:case11pct_undef_def} in both the undeformed and deformed configurations.
The resulting displacement of the actuated arm is in this case, as expected, somewhat smaller, with  $T_3\!=\!\SI{-1.625}{\milli\meter}\!=\!-0.1083 R$ and $\theta_2\!=\!\SI{0.18016}{\radian}\!=\!\SI{10.322}{\degree}$.
Table~\ref{tab:design_perfomance_comparison} presents a detailed comparison of the performance of the two designs, in terms of rigid arm rotation, spring elongation, and the main objective value $C_0$.

\begin{figure}[b!]
\centering
\begin{overpic}[height=0.43\linewidth,trim=0mm 3mm 0mm -9mm, abs]
               {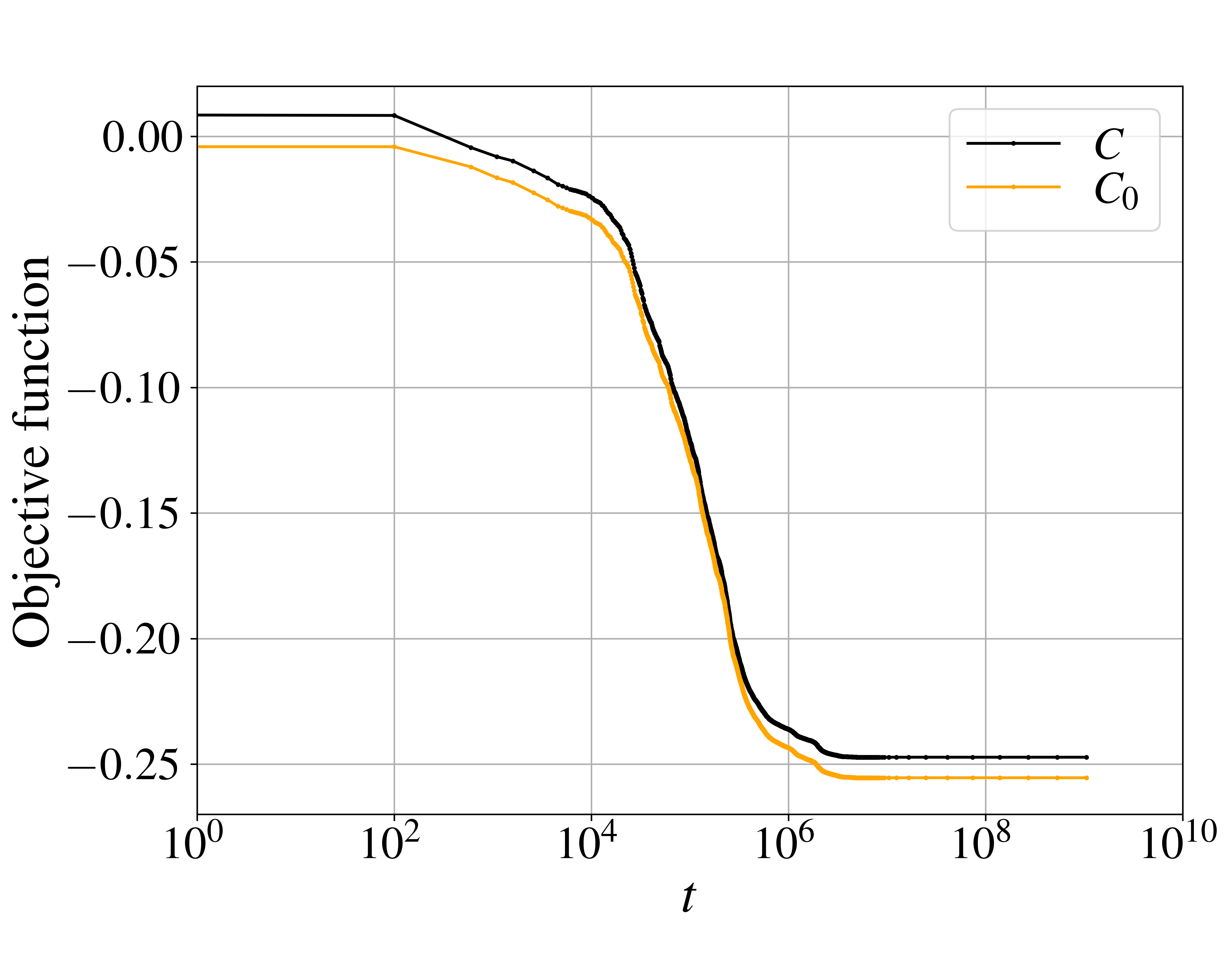}
\put(93,189){$\sqrt{2\Psi_\mathrm{lim}/E}=14\%$}
\end{overpic}
\begin{overpic}[height=0.43\linewidth,trim=31mm 3mm 1mm -9mm,clip, abs]
               {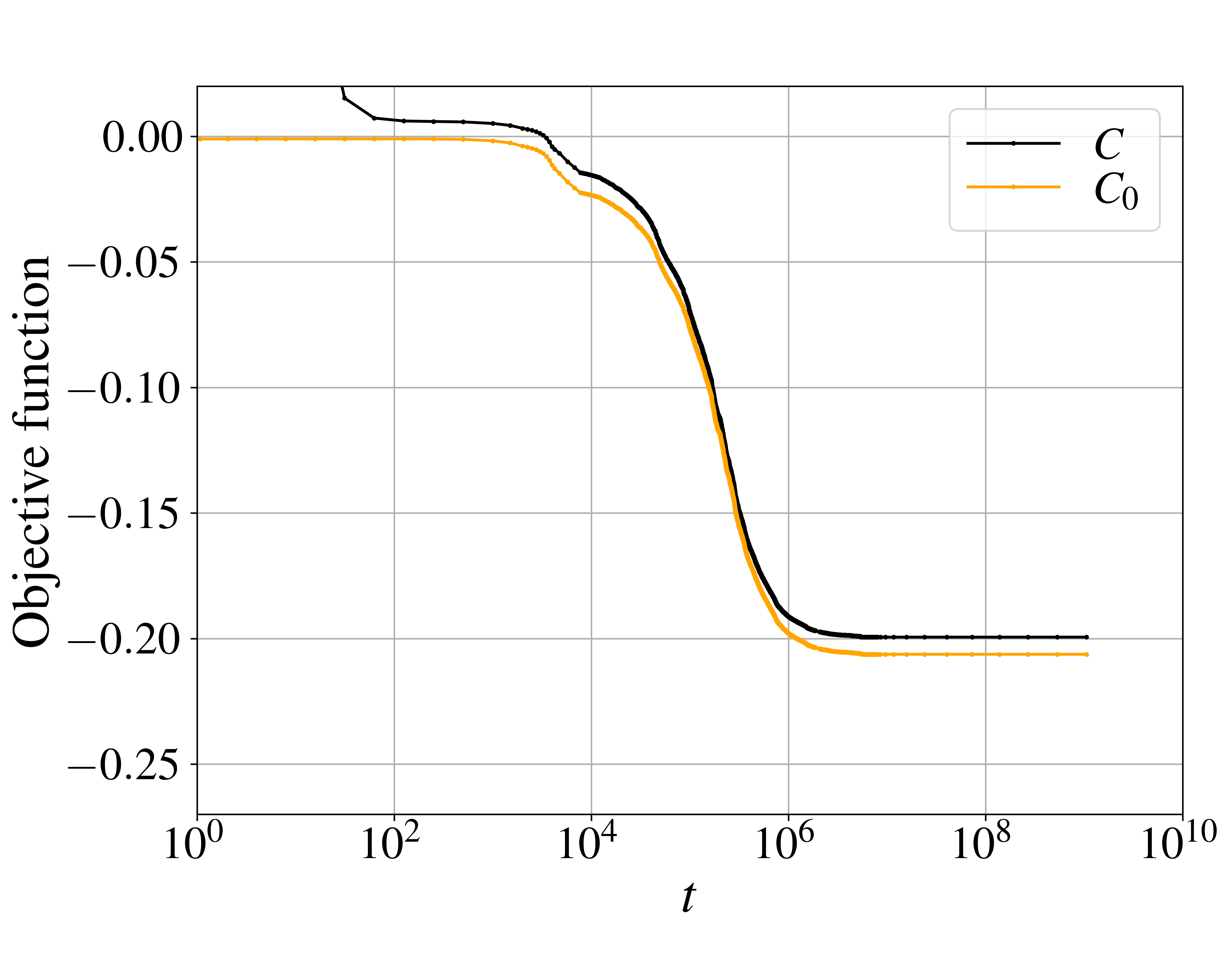}
\put(60,189){$\sqrt{2\Psi_\mathrm{lim}/E}=11\%$}
\put(0,25){\textcolor{white}{\rule{5pt}{157pt}}}
\end{overpic}
\caption{Convergence history for the two optimization cases.}
\label{fig:convergence_history}
\end{figure}

Additional information about the optimization history is provided in the diagrams of Figure~\ref{fig:convergence_history}.
The two diagrams show the evolution of the total objective $C$ and the main objective $C_0$ over optimization pseudo-time, respectively for the two distinct strain energy density limits the optimization algorithm was run with.
In any case, the total objective decreases monotonically as expected with the type of employed optimization algorithm.
It is also clear from the diagrams that the main objective $C_0$, representing the bending response of the actuated arm, is the dominant term in $C$.
Only in the beginning of the second optimization case with the low $\Psi_\mathrm{lim}\!=\!\tfrac{1}{2}0.11^2E$, the term $C_\Psi$ is temporarily dominating $C$, because the initial design did not satisfy the imposed strain energy density constraint.
The ultimate objective value for the case with the stricter strain energy density constraint is as expected higher than in the first case.
For more details about the evolution of the remaining contributions $C_A$, $C_i$, $C_\Psi$, and $C_p$ to the total objective, the reader is referred to Appendix~\ref{appendix_CA_Ci_CPsi_Cp}.

\section{Experimental Methodology}

The basic dimensional analysis that allowed us to express all dimensional parameters in Table~\ref{tab:model_params} as a function of one length parameter $R$ and one stiffness parameter $E$, allows us also to easily rescale the system in terms of both size and stiffness.
To accommodate practical aspects of the manufacturing, all experimental results to be presented in the remaining of this work were obtained with an upscaled actuator radius $\bar{R}\!=\!1.5 R\!=\!\SI{22.5}{\milli\meter}$ and a decreased initial Young's modulus for the elastomer equal to $\bar{E}\!=\!\SI{1}{\newton\per\milli\meter\squared}$, i.e. with a 36.55\% scaling of the original stiffness.

\subsection{Manufacturing of optimized designs}

\begin{figure}[b!]
  \tikzstyle{block}=[rectangle,
                     rounded corners,
                     fill=red!20,
                     minimum width=2.1cm,
                     minimum height=0.5cm,
                     text centered,
                     text width=2cm,
                     draw=black]
  \tikzstyle{arrow} = [thick,->,>=stealth]

  \begin{center}
  \begin{tikzpicture}[node distance=4.2cm]
    \node (levset) [block,text width=3.7cm]
      {Level-set Smoothing \& Thresholding};
    \node (slice) [block, right of=levset]
      {Supports \& Slicing};
    \node (print) [block, right of=slice]
      {SLA 3D printing};
    \node (wash) [block, right of=print,text width=3.6cm]
      {Washing, Curing\\ \& Support Removal};
  
    \draw [arrow] (levset) -- (slice);
    \draw [arrow] (slice) -- (print);
    \draw [arrow] (print) -- (wash);

    \node (levsetvis) [below of=levset,yshift=1.5cm]
      {\includegraphics[height=0.2\textwidth]
       {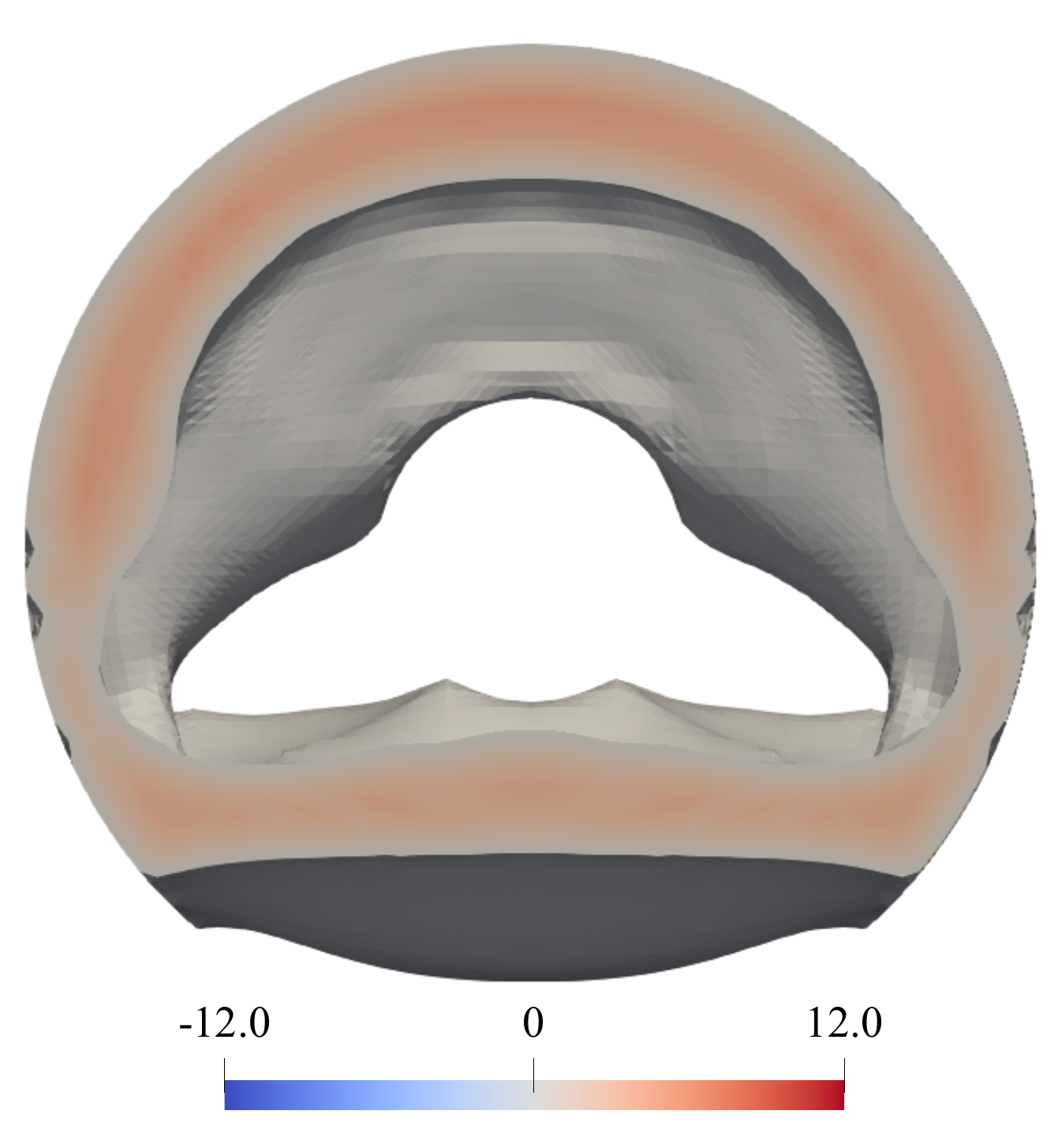}};
    \node (slicevis) [below of=slice,yshift=1.5cm]
      {\includegraphics[height=0.2\textwidth]
       {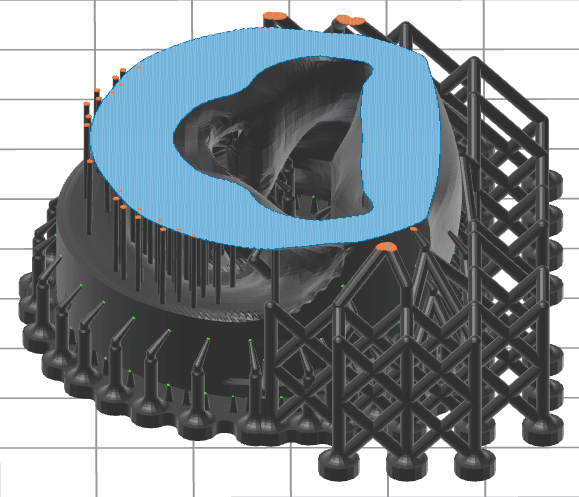}};
    \node (printvis) [below of=print,yshift=1.5cm]
      {\includegraphics[height=0.2\textwidth]
       {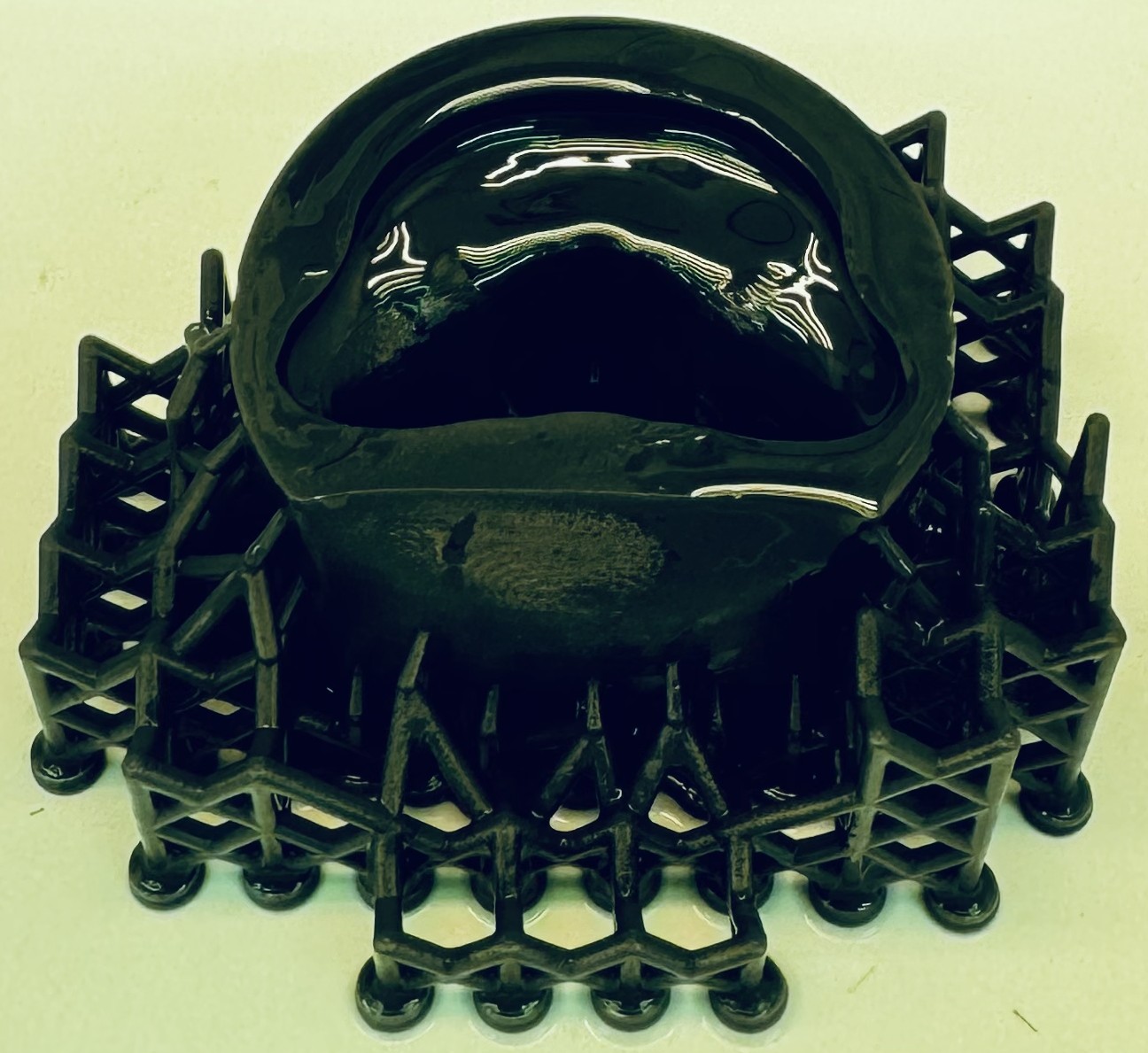}};
    \node (washvis) [below of=wash,yshift=1.5cm]
      {\includegraphics[height=0.2\textwidth]
       {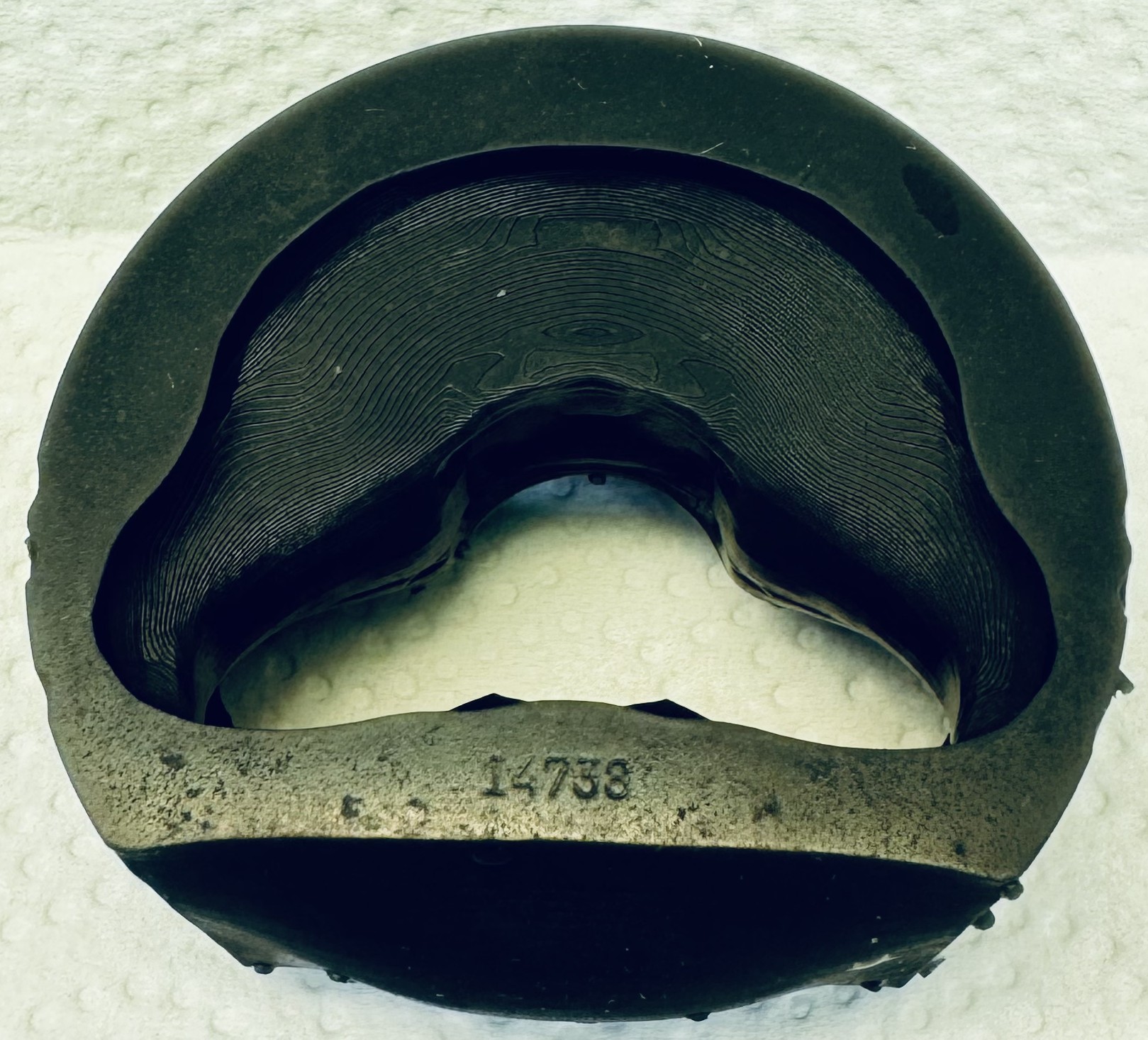}};
  \end{tikzpicture}
  \end{center}
  \caption{Flowchart for SLA 3D printing through Formlabs printers.}
  \label{fig:manufacturing_flowchart}
\end{figure}
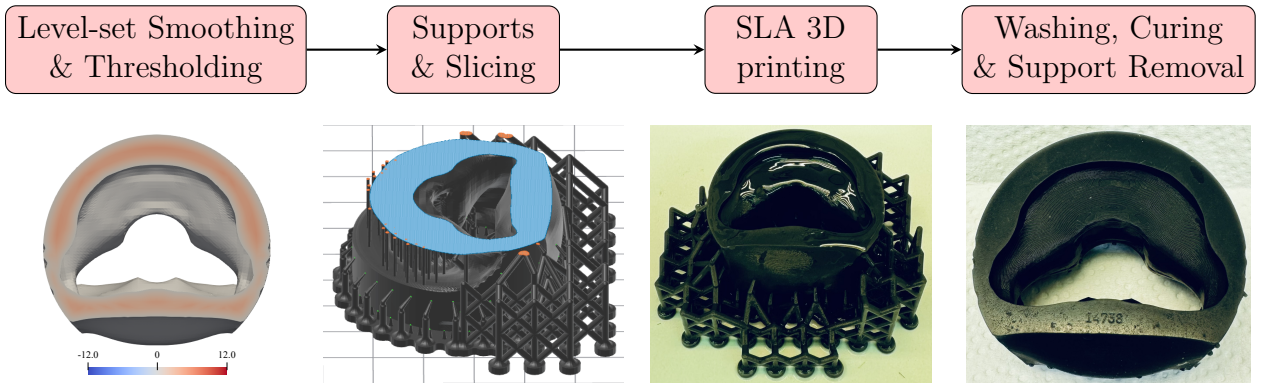

The two optimized actuator designs, presented in the previous section, were 3D printed in the resin based stereolithography (SLA) printer Form 3 by Formlabs.
The entire procedure from obtaining the converged design field $\chi$ to receiving the final manufactured component, is illustrated in Figure~\ref{fig:manufacturing_flowchart}.

In a first step, in order to mitigate the impact of the relatively coarse discretization on the design, the obtained level-set field $\chi$ is mapped onto a finer mesh and it is then filtered with the Helmholtz-type PDE filter \cite{2011La-Si}.
The finer mesh has twice as fine discretization in all three directions, and the radius of the applied Helmholtz filter is $r\!=\!\SI{0.1}{\milli\meter}$.
After obtaining the filtered level-set field $\bar{\chi}$ on the refined mesh, by solving the equation
\begin{equation}\label{eq:chi_pde_filter}
\int_\Omega (\bar{\chi}-\chi) \delta\bar{\chi} + r^2\,\nabla \bar{\chi} \cdot \nabla\delta \bar{\chi} ~d\Omega = 0,
\end{equation}
the design was further processed in ParaView.
Due to issues in ParaView with slicing of higher order elements, the imported dataset was first passed through the Tetrahedralize filter.
Then the Transform filter was used to mirror the tetrahedralized meshed, and the two symmetric halves were merged using the GroupDataset and MergeBlocks filters.
Finally, the Clip filter was used to retain only the solid portion of the domain for $\bar{\chi}>0$, see first image in Figure~\ref{fig:manufacturing_flowchart}, and the surface of the solid was extracted using the ExtractSurface filter, and exported in an STL file.

Each exported STL file for the two actuator designs was imported into the PreForm\textsuperscript{\tiny\textregistered} software by Formlabs to add the necessary support structure and perform the slicing.
At the same time, the two geometries were upscaled with the aforementioned factor $\bar{R}/R\!=\!1.5$.
Upscaling of the geometry allows for an improved manufacturing accuracy and easier handling of the experimental setup in general.
The slicing process converted the 3D geometry into $270$ printing layers with a thickness of \SI{0.1}{\milli\meter} per layer.
The support structures were generated through the automatic support feature in the Preform\textsuperscript{\tiny\textregistered} software.
The second image in Figure~\ref{fig:manufacturing_flowchart} shows a snapshot of the support structure and the printed component at slicing layer $212$ for the actuator design from Figure~\ref{fig:case14pct_3Dviews}.

Both optimized designs were printed in the SLA printer Form 3 by Formlabs using a silicone resin available by the printer manufacturer under the commercial name Silicone 40A~V1.
According to the manufacturer, this material has as a shore hardness of 40A, an ultimate tensile strength of \SI{5}{\newton\per\milli\meter\squared}, an ultimate elongation of  230\%, and an initial Young's modulus in the order of \SI{1}{\newton\per\milli\meter\squared}.
The latter value is used for scaling of the optimization results to a realistic elastomer stiffness level with the aforementioned factor $\bar{E}/E\!=\!0.3655$.
Each print requires approximately six hours of print time, followed by a washing process, in which the printed part is immersed in isopropyl alcohol inside an ultrasonic cleaner operating at a frequency of \SI{37}{\kilo\hertz}. 
This cleaning process was repeated two to three times before manually removing all support structures with a cutter and letting the actuator to air-dry. 
Finally, the printed actuator was submerged in lukewarm water and post-cured under UV light at \SI{60}{\degreeCelsius} for 45 minutes.
The last image in Figure~\ref{fig:manufacturing_flowchart} shows 
the final cured actuator for the optimized design from Figure~\ref{fig:case14pct_3Dviews}.

\subsection{Testing device}

\begin{figure}[t!]
\begin{overpic}[width=\linewidth,trim=0cm 0cm -20cm -5cm,clip]
               {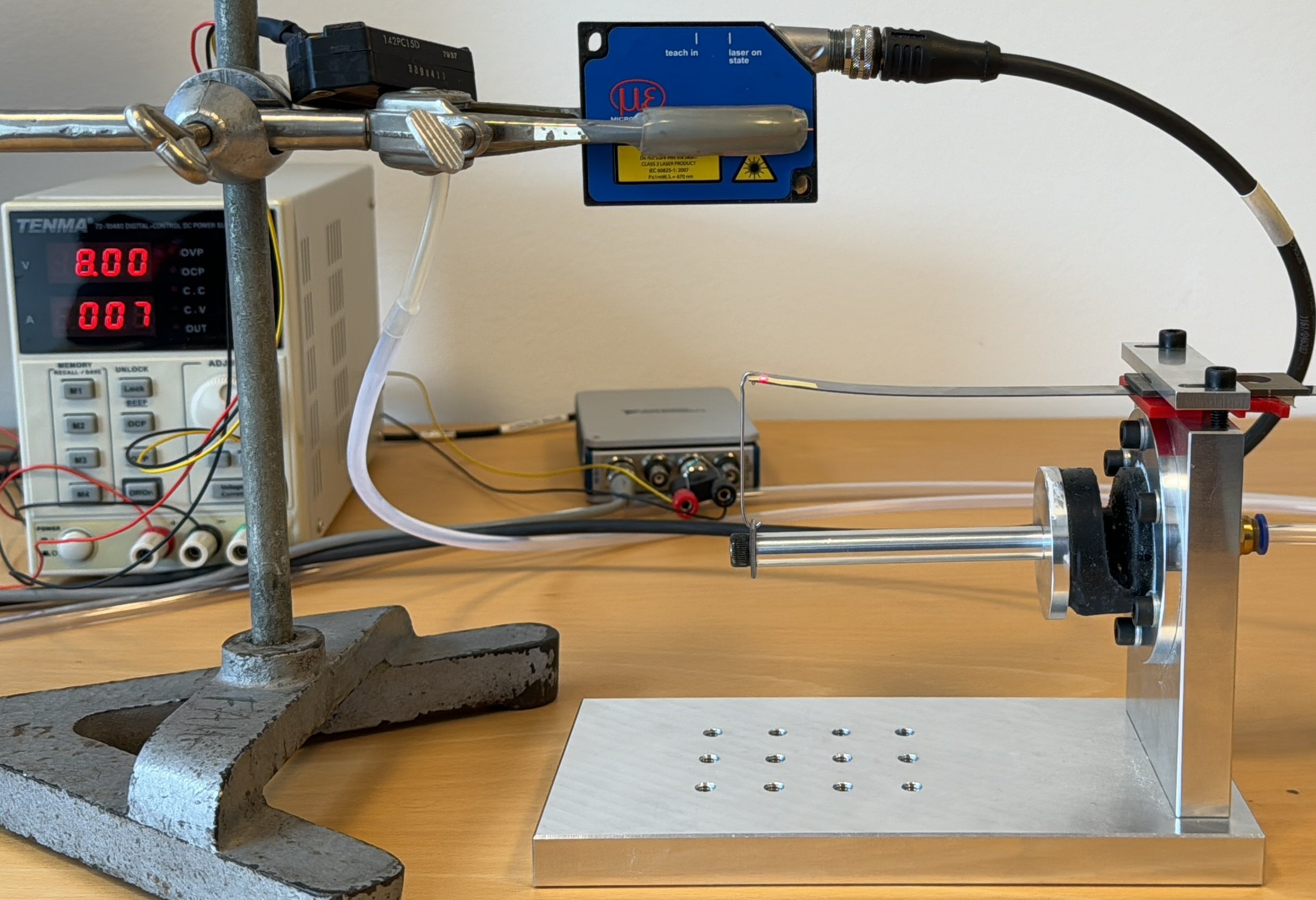}
  \put(80.5,3)
  {\includegraphics[width=0.2\textwidth,trim=0cm 0cm 0cm 0cm,clip]
   {experimental/manufacturing_Wlim14pct_14738/3D_printed_14738.JPG}
  }
  \put(81,1){3D-printed design}
  \put(81.5,13){\color{magenta}\vector(-2,1){11}}
  \put(81.5,26){\color{magenta}\vector(-1,-1){3}}
  \put(82,26){pressure supply}
  \put(81.5,52){\color{magenta}\vector(-2,-1){42}}
  \put(82,53){data acquisition}
  \put(82,50){board (DAQ NI)}
  \put(81.5,45){\color{magenta}\vector(-3,-2){21}}
  \put(82,45){cantilever spring}
  \put(81.5,38){\color{magenta}\vector(-3,-2){17}}
  \put(81.5,38){\color{magenta}\vector(-1,-1){10}}
  \put(82,38){back plates}
  \put(42,56){\color{magenta}\vector(0,-1){3}}
  \put(38,57){displacement sensor}
  \put(22,56){\color{magenta}\vector(0,-1){3}}
  \put(18,57){pressure sensor}
  \put(4,56){\color{magenta}\vector(0,-1){13}}
  \put(0,57){power supply}
\end{overpic}
\caption{Experimental testing device and instrumentation.}
\label{fig:exper_setup}
\end{figure}

The setup of the optimization problem from Figure~\ref{fig:init_design} was replicated experimentally in the device shown in Figure~\ref{fig:exper_setup}.
To achieve exactly the same boundary conditions for the elastomer component, the printed actuator was affixed with a strong adhesive on both sides to circular discs made of aluminum, as shown in the figure.
The gluing process needs to provide a connection which is not only strong enough but also airtight to prevent any leakage upon pressurization.
To achieve this, the aluminum surfaces were treated with the commercially available primer DOWSIL\texttrademark~1200~OS and left to cure for an hour before gluing the elastomer onto the treated aluminum surface with the silicone adhesive DOWSIL\texttrademark~734~Clear.
Finally the interface was allowed to cure for at least 24 hours to ensure the development of a reliable and leak-proof bond before used in testing.

The aluminum disc on the fixated side of the actuator is attached to the vertical plate of the aluminum frame of the testing device with screws and sealed with an O-ring.
Pressurized air is provided through a hole in the center of the disc.
The aluminum disc glued on the opposite side of the actuator is connected to a cylindrical tube made of aluminum, with a total length of $\bar{L}_\mathrm{arm}\!=\!4\bar{R}\!=\!\SI{90}{\milli\meter}$, in accordance with the data from Table~\ref{tab:model_params}.

To obtain a spring stiffness equivalent to the numerical model, a linear spring is required that is scaled to the actual dimensions and material stiffness in the testing device.
According to Table~\ref{tab:model_params}, the stiffness of the spring should be equal to
$\bar{k}_\mathrm{sp}\!=\!1.6\cdot10^{-2} \bar{E}\,\bar{R}^3/\bar{L}_\mathrm{arm}^2$.
This is only for one half of the actuator, so this value needs to be multiplied by 2 for the entire actuator, resulting in a stiffness value of \SI{0.045}{\newton\per\milli\meter} for the actual spring in the testing device.
To implement this target stiffness value, a cantilever spring was used made of a \SI{0.5}{\milli\meter} thin and \SI{12.7}{\milli\meter} wide stainless steel plate.
Assuming a Young's modulus of \SI{200}{\giga\pascal} for the spring material, the desired stiffness could be obtained with a free length of \SI{120.8}{\milli\meter}.
Rather than relying on such an assumption, the length of the cantilever spring was directly calibrated experimentally to provide the desired stiffness, see Appendix~\ref{appendix_spring_calib}, resulting in a free length of~\SI{113}{\milli\meter}.
The end of the cantilever spring was connected to the end of the actuated arm with a steel wire.

The setup is equipped with two sensors, a displacement sensor from Micro-Epsilon with the model name ILD 1402-50, and a pressure sensor from Honeywell with the model name 142PC15D.
The displacement sensor is from the Micro-Epsilon optoNCDT 1402 series that relies on laser triangulation to measure changes in distance, and covers the distance range \mbox{45-95}\,\si{\milli\meter}, with a sensitivity of \SI{12.5}{\milli\meter\per\volt}.
This sensor was mounted at approximately \SI{50}{\milli\meter} above the cantilever spring to measure the vertical deflection of the end of the spring which is connected to the end of the actuated arm.
The Honeywell pressure sensor measures differential pressure between two ports P1 and P2, within the range 0-\SI{15}{\PSI}, with a theoretical sensitivity of \SI{3}{\PSI/\volt}.
However, recalibration of this sensor with a water column attached to port P2 resulted in a sensitivity of \SI{2.78}{\PSI\per\volt}, i.e. \SI{19.17}{\kilo\pascal\per\volt}, which was used for all measurements.

Voltage signals from the two sensors were recorded through two out of the four channels of a National Instruments NI-9215 voltage input module covering the measuring range of $\pm10$\,\si{\volt}.
The module was connected to a PC through the USB interface of a cDAQ-9171 chassis by National Instruments.
A very simple Python script, based on the Python library NI-DAQmx, was used to record the voltage signals of both sensors during the test, at intervals of \SI{0.2}{\second} and 100 samples per read.

The pressurization of the system was done manually by means of a pressure regulator by FESTO with the model name LR-1/4-DB-7-MINI.
The pressure regulator, which is not included in Figure~\ref{fig:exper_setup}, was connected between the inlet of the actuator shown at the right hand side of the figure, and an accumulator of pressurized air with sufficiently high pressure and volume.
The target pressure which is equivalent to the input pressure used in the optimization cases, can be computed according to Table~\ref{tab:model_params} as $\bar{p}_\mathrm{in}\!=\!2\!\cdot\!10^{-2}\bar{E}$, resulting in the value of \SI{20}{\kilo\pascal}.
During all tests, discussed in the next section, the inlet pressure was manually increased up to a point exceeding the target pressure of \SI{20}{\kilo\pascal} by a safe margin, and then the pressure was slowly decreased again, before any subsequent repetition.

\section{Experimental Results and Comparisons}

\begin{figure}[b!]
\centering
\begin{overpic}[width=\linewidth,trim=-8cm 7mm -8cm -22.2cm,clip]
               {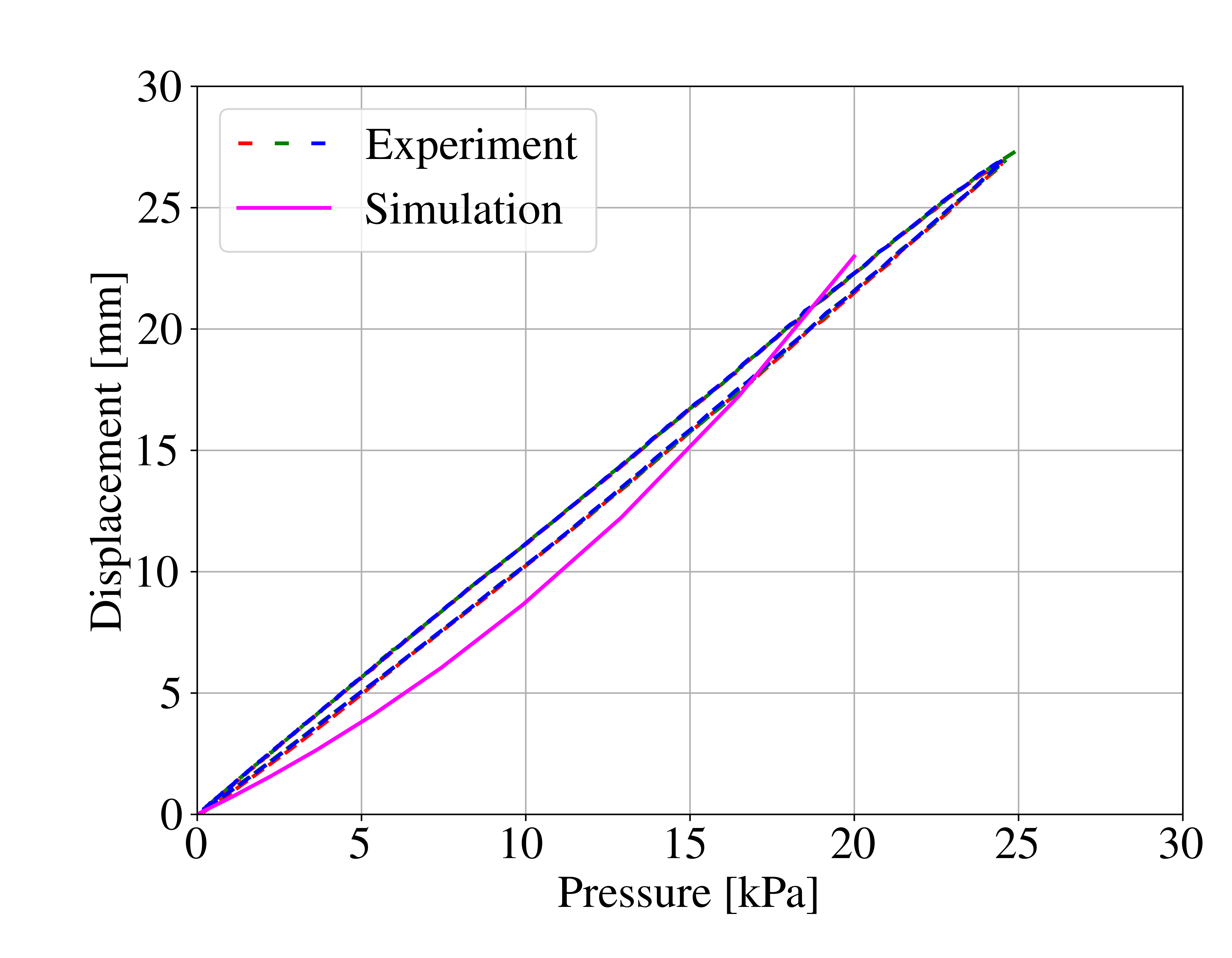}
\put(1,42){\reflectbox{
           \includegraphics[width=0.48\linewidth,trim=40cm 10cm 0cm 21cm,clip]
                           {experimental/undef_14738.JPG}}}
\put(51,42){\reflectbox{
           \includegraphics[width=0.48\linewidth,trim=40cm 10cm 0cm 21cm,clip]
                           {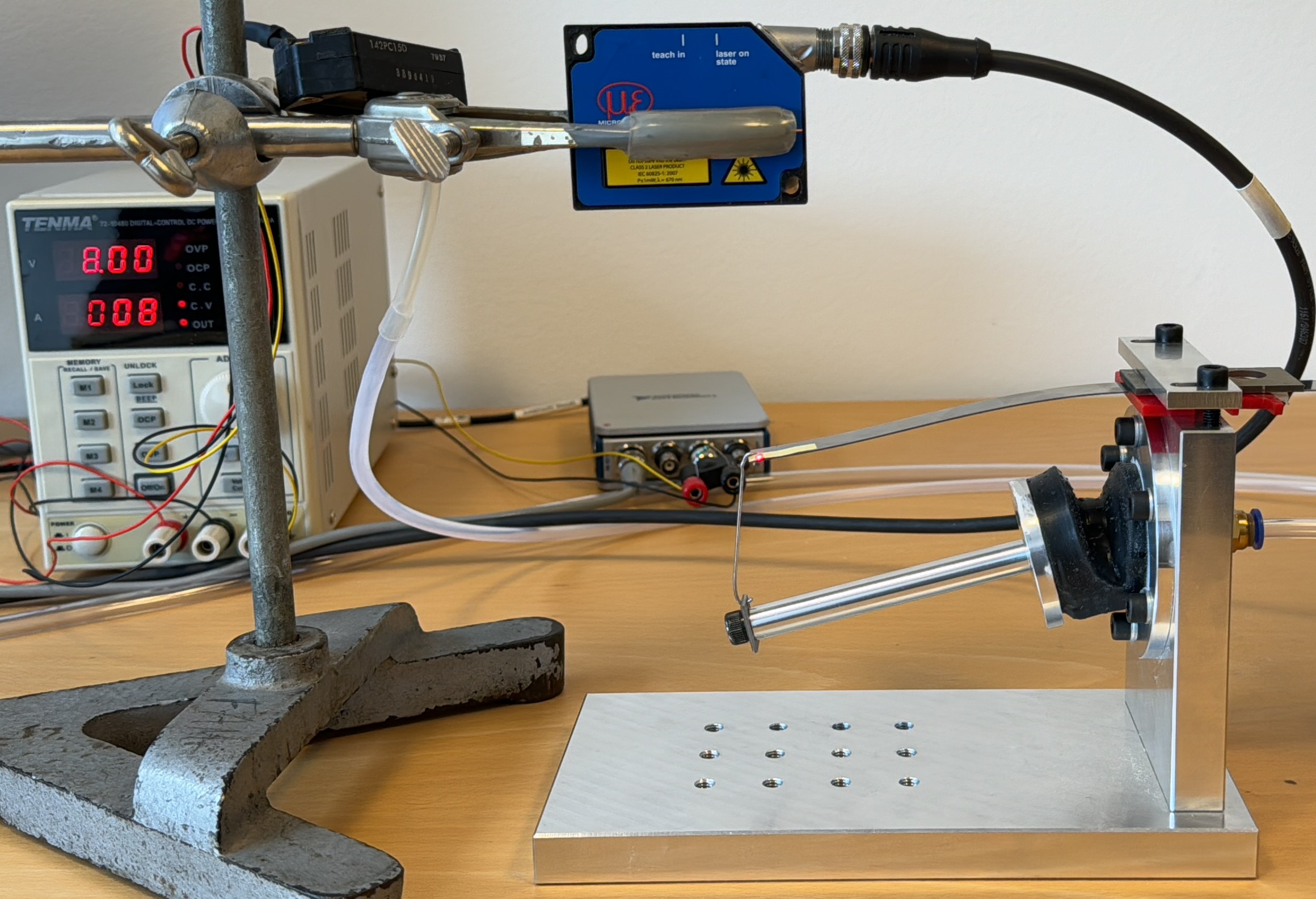}}}
\put(10,70.9){\includegraphics[width=0.37\linewidth,trim=0cm 0cm 0cm 0cm,clip]
                           {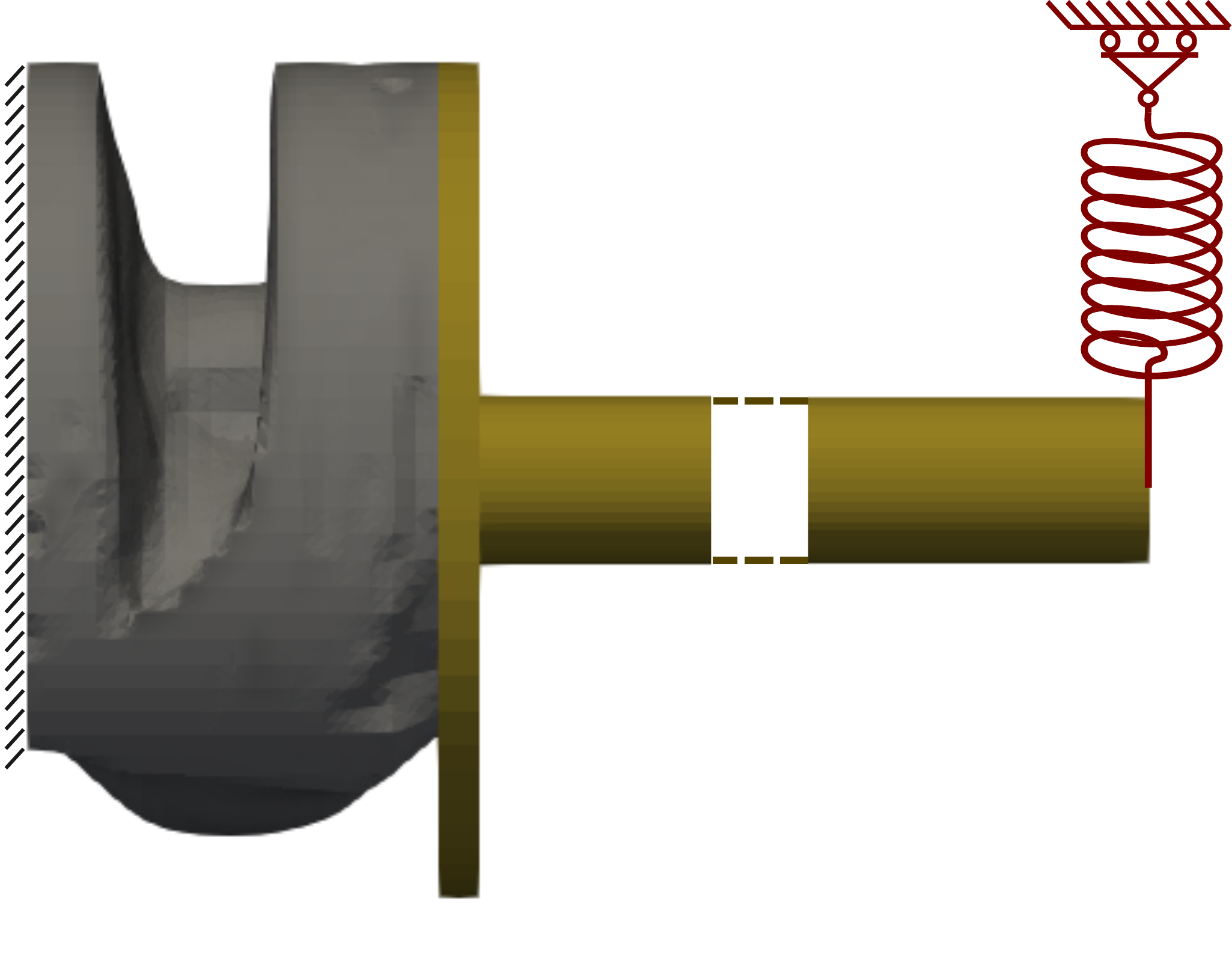}}
\put(60,70.2){\includegraphics[width=0.37\linewidth,trim=0cm 0cm 0cm 0cm,clip]
                           {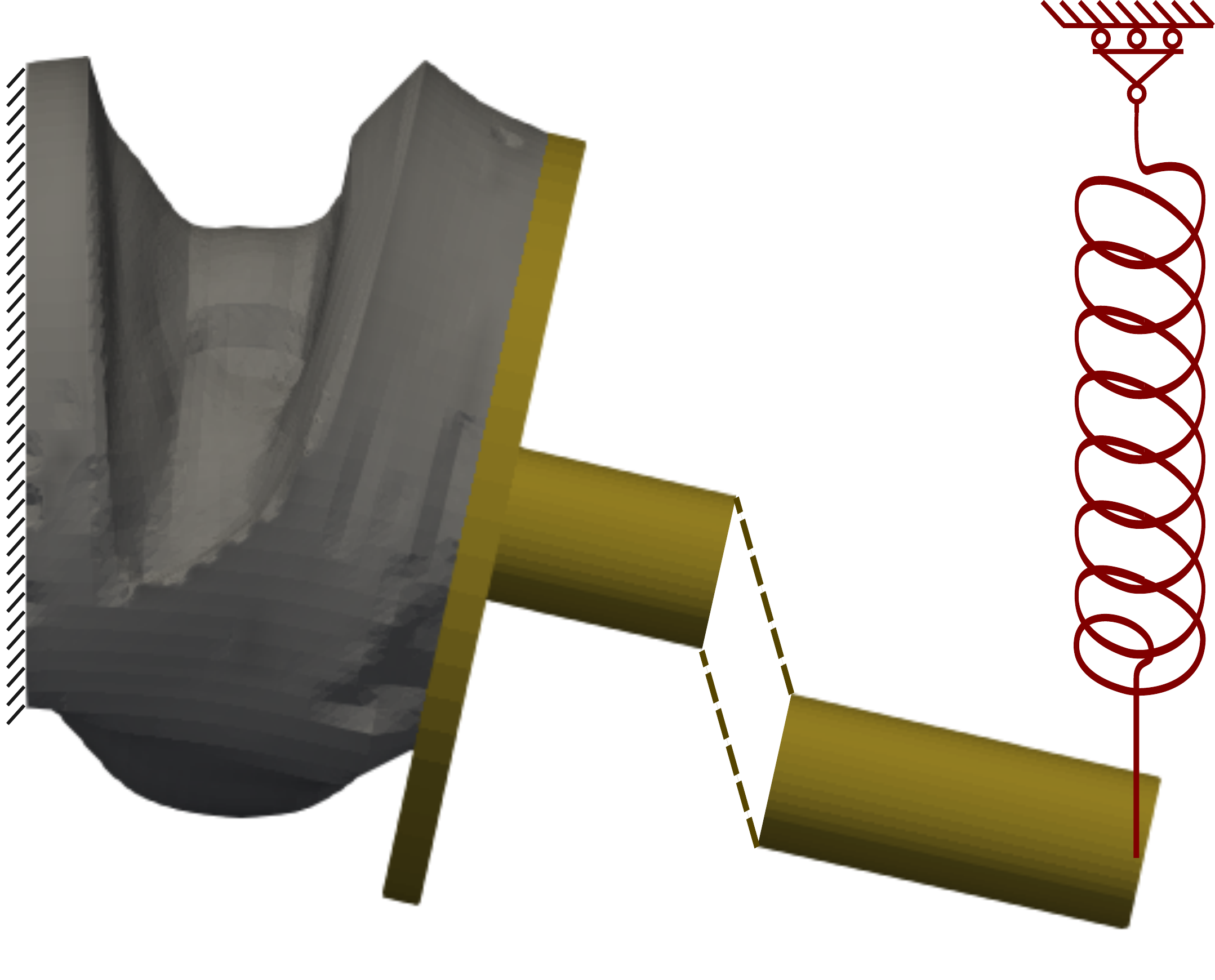}}
\put(0,98){a)}
\put(0,71){b)}
\put(0,39){c)}
\end{overpic}
\caption{a) Numerically obtained actuator design for $\sqrt{2\Psi_\mathrm{lim}/E}\!=\!14\%$ in its undeformed and deformed configurations, b) corresponding physical realization before and after pressurization, and c) comparison of pressure-displacement data between numerical simulation and experiment.}
\label{fig:exper_case14pct}
\end{figure}

The testing results for the main optimized design of this work, obtained for $\sqrt{2\Psi_\mathrm{lim}/E}\!=\!14\%$, are presented in Figure~\ref{fig:exper_case14pct}.
Figures~\ref{fig:exper_case14pct}a and~\ref{fig:exper_case14pct}b respectively show the numerical and physical systems in both undeformed and deformed configurations.
The three overlapping dashed curves in the pressure-displacement graph in Figure~\ref{fig:exper_case14pct}c represent three repetitions of the pressurization-depressurization cycle, which extend beyond the target pressure of \SI{20}{\kilo\pascal}.
The solid line represents the predicted response from the simulation, scaled of course according to the radius $\bar{R}$ and Young's modulus $\bar{E}$ relevant for the experiment.
The experimental results exhibit rather low hysteresis for an elastomer, and are very reproducible even with the manual control of the provided pressure.
The same loading and unloading curves are reproduced despite the inevitable small variations in the loading rate from repetition to repetition.

The main discrepancy between the numerical prediction and the experimental result concerns the much more linear response observed in the experiment compared to the simulation.
However, for any practical purposes, the experimentally reproduced behavior actually satisfies the goal of the optimization.
For the prescribed actuation pressure and resistance from the environment, the actuator produces the response that it was optimized for.
Given the large uncertainty in the exact constitutive law of the 3D printed elastomer, this result is very encouraging for future applications of topology optimization in designing soft actuators.

\begin{figure}[b!]
\centering
%\frame{
\begin{overpic}[width=\linewidth,trim=-8cm 7mm -8cm -12.8cm,clip]
               {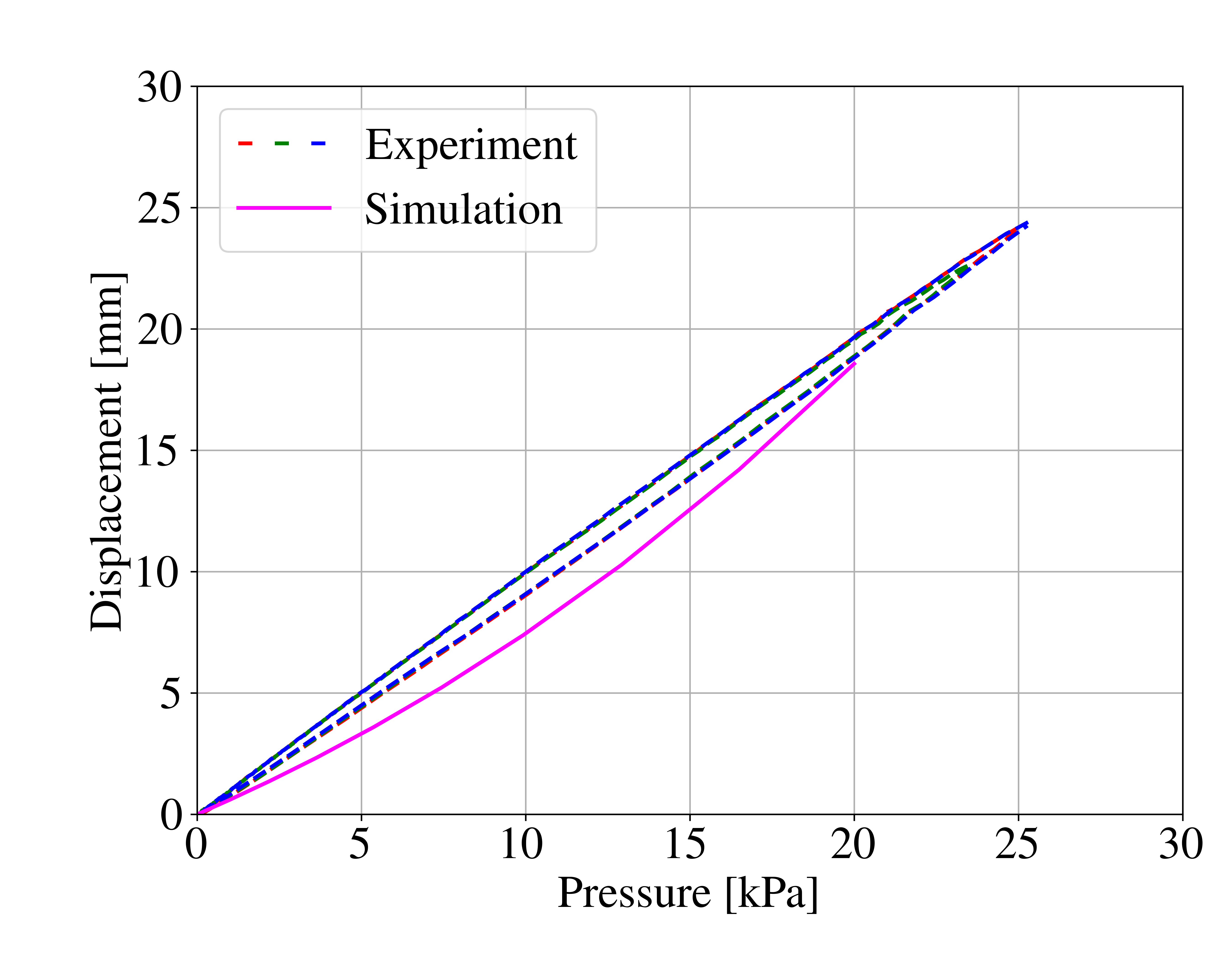}
\put(1,42){\reflectbox{
           \includegraphics[width=0.48\linewidth,trim=40cm 10cm 1cm 22cm,clip]
                           {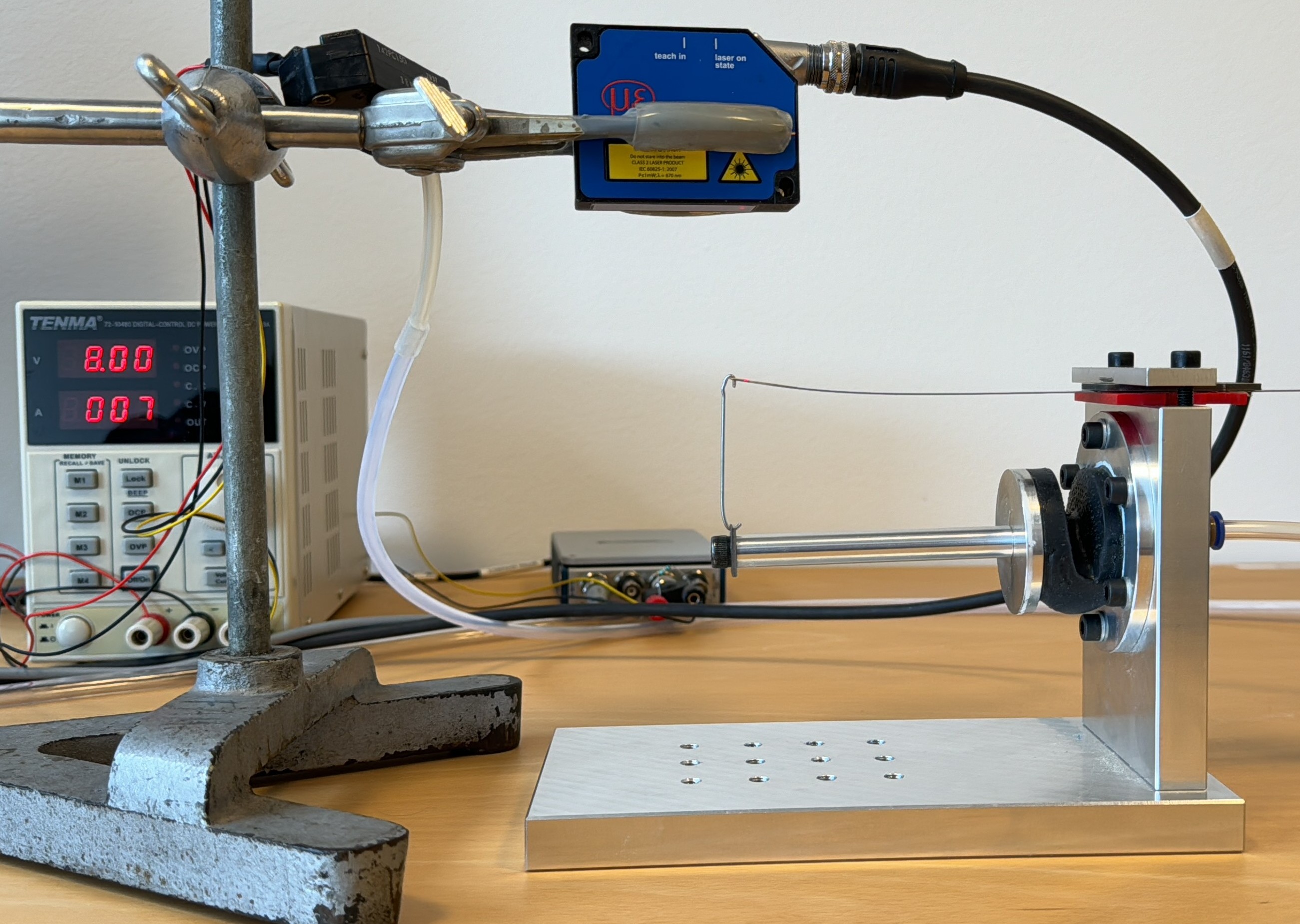}}}
\put(51,42){\reflectbox{
           \includegraphics[width=0.48\linewidth,trim=40cm 11cm 0cm 22cm,clip]
                           {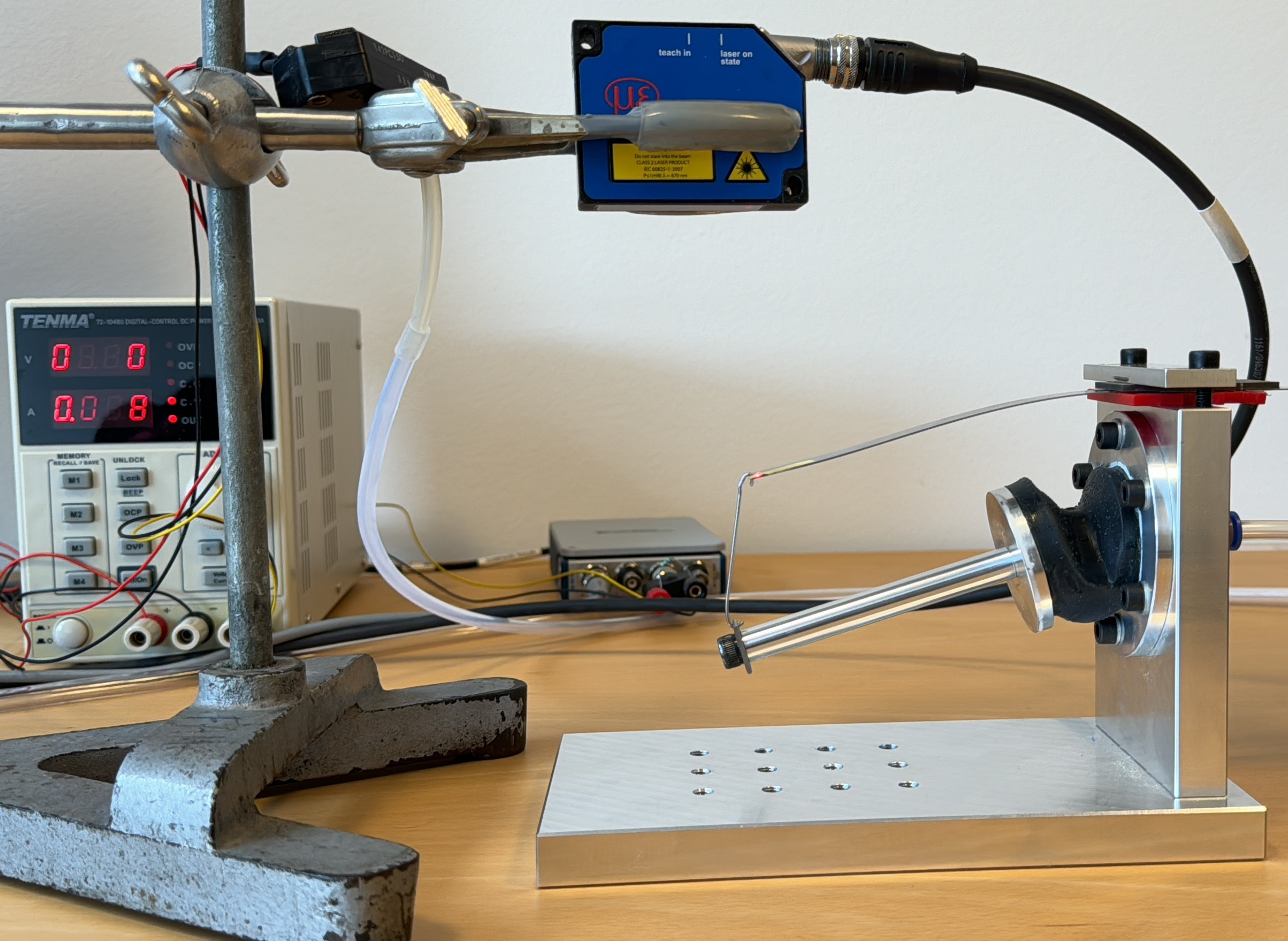}}}
\put(0,74){a)}
\put(0,39){b)}
\end{overpic}
\caption{a) Physical realization of the actuator design obtained for $\sqrt{2\Psi_\mathrm{lim}/E}\!=\!11\%$  before and after pressurization, and b) comparison of pressure-displacement data between numerical simulation and experiment.}
\label{fig:exper_case11pct}
\end{figure}

Visual comparison of the deformed actuator between simulation and experiment, provides some additional confidence in the 3D printed component being close to the optimized one both in geometry and mechanical response.
Of course this is not a direct validation of how well the strain energy density constraint, set in the optimization, is actually satisfied, just a qualitative assessment.

Figure~\ref{fig:exper_case11pct} presents the testing results for the second optimized design, obtained with the stricter strain energy density constraint.
Results are similar to the first case, just as expected with a lower level of displacements for the same actuation pressure, compared to the first case.
At the target pressure of \SI{20}{\kilo\pascal}, that the actuator was optimized for, the experimentally obtained displacement is even slightly higher than the numerical prediction.
Otherwise, again, the experimental response is almost linear, while the numerical pressure-displacement relationship is clearly nonlinear.

\section{Concluding Remarks}
This work has presented a comprehensive treatment of the computational design of soft pneumatic actuators, from fundamental numerical modeling to practical implementation of a physical device.
By tracing the full path from the theoretical formulation to manufacturing and in-service performance, we have demonstrated the viability of computational design of this type of actuators, based on topology optimization and SLA 3D printing.

An essential feature of the numerical framework of this work is that it both accounts for the very large deformations in the actuator as well as for a certain resistance from the environment, as an inherent part of the optimization.
This is in contrast to conventional approaches that often aim at just producing motion in unloaded conditions, and often ignore the effect of large deformations during the optimization.
It is also important to highlight the incorporation of an allowable strain constraint.
This constraint is mainly meant to prevent excessive stretching that could lead to mechanical failure, and in particular to avoid designs that rely on ballooning.
At the same time, in a real world with big uncertainties in elastomer material behavior at very large strains, designing specifically for lower strain levels enhances the robustness and reliability of the design.

Finally, to ensure the reproducibility and comparability of our findings, we have introduced a well-defined benchmark setup for a soft actuator module and discussed practical aspects of its construction and instrumentation.
The testing of our designs in the test bench revealed a certain deviation from the simulation, in terms of how linear the pressure-displacement response is, but an overall good agreement with the numerically predicted performance of the 3D-printed actuator.
A priority for future work is to use specific hyperelastic models during the simulations and optimization, fitted to the specific elastomer used in the printing, rather than a generic hyperelastic model.

The employed numerical framework has two main limitations worth noting.
First, the optimization accounts only for the pressurized deformed configuration, without examining the transition from the non-pressurized to the pressurized state.
Post-simulations of the designs obtained in this work confirmed that the optimized pressurized state is attainable, but this is not guaranteed in general for highly nonlinear systems.
Second, the framework carries a high computational cost.
The current second-order optimization scheme is capable of reaching the exact optimality point, but at the expense of costly design updates throughout the entire optimization history.
A more balanced approach would be to use a first-order scheme for the main part of the optimization, switching to second-order only when approaching the optimality point.

\section*{Acknowledgments}
This work was supported by the research grant VIL50407 from VILLUM FONDEN.

\section*{Appendices}
\renewcommand{\thesubsection}{\Alph{subsection}}

\subsection{Pressurization source}
\label{appendix_Qin}
The pressure source intensity term $Q_{\mathrm{in}}$ is illustrated in Figure~\ref{fig:init_design} only visually.
In order to use it in Eq.~\eqref{eq:Darcy_flow}, its exact mathematical definition is required.
It is defined as
\begin{equation}
Q_{\mathrm{in}}(\XX)
=100\,Q_s
 \left(1-\sin^2\left(\dfrac{\pi}{2}
                     \min\left(1,\left<\dfrac{10\,r(\XX)}{R}
                                       -\dfrac{3}{2}\right>
                         \right)
                \right)
 \right),
\end{equation}
with
\begin{equation}\label{eq:radius}
  r(\XX)=\sqrt{X_2^2 + X_3^2}.
\end{equation}
representing the radial coordinate of any point $\XX$ within the computational domain $\Omega$.
According to this definition, $Q_{\mathrm{in}}$ is zero outside a core cylinder with radius $0.25R$ and it transitions fast to $100\,Q_s$, in the interior of this region.

\subsection{Passive void region}
\label{appendix_Cx}
A prerequisite for the following definitions is the so called LogSumExp smooth approximation of the min function.
It is defined, including a proper scaling of its arguments, as
\begin{equation}\label{eq:maxLSE}
\min_\mathrm{LSE}(a,b) = -\frac{4}{L_i} \mathrm{log}(e^{-a\,L_i/4} + e^{-b\,L_i/4}).
\end{equation}
The passive void region penalization function $C_\chi$, used in Eq.~\eqref{eq:aug_objfunc}, can then be defined as
\vspace{-5pt}
\begin{equation}\label{eq:Cx}
  C_\chi(\chi)=\dfrac{10^5}{2}
               \left<\chi-\dfrac{8}{L_i}
                          \min\!\left(R\!-\!r(\XX),~
                                      -\min_{LSE}\!\left(r(\XX),-r(\XX)\right)
                                      -0.3 R\right)
               \right>^2.
\end{equation}
The role of this term in Eq.~\eqref{eq:aug_objfunc} is to enforce a cylindrical passive void region with a radius of~$0.3R$ in the core of the computational domain.
This is a bit larger than the pressurized region of radius $0.25R$, shown in Figure~\ref{fig:init_design}, and defined in Appendix~\ref{appendix_Qin}.

\subsection{Initial design}
\label{appendix_init_design}

The exact mathematical definition of the initial design illustrated in Figure~\ref{fig:init_design}, is

\vspace{-8pt}
\begin{equation}\label{eq:chi_0}
\begin{aligned}
\left.\chi\right|_{t=0}=
  \min_\mathrm{LSE}
  \Biggl(&\dfrac{8}{3 L_i}\min_\mathrm{LSE}
                          \Bigl(0.95 R\!-\!r(\XX),~ r(\XX)\!-\!0.45 R\Bigr)
          -0.1\left<X_3\right>,\\[-2pt]
         &\dfrac{8}{L_i}\min_\mathrm{LSE}\Bigl(R\!-\!r(\XX),~ r(\XX)\!-\!0.3 R\Bigr)
  \Biggr).
\end{aligned}
\end{equation}
It resembles a tube with inner radius of $0.45R$ and outer radius of $0.95R$.
However, the term $0.1\left<X_3 \right>$ decreases slightly the wall thickness in the positive $X_3$ half-space, in order to introduce some initial asymmetry.
Moreover, the last part of the expression ensures that the initial design does not violate the passive void region constraint defined in Eq.~\eqref{eq:Cx}.

\subsection{Evolution of penalization terms in the objective}
\label{appendix_CA_Ci_CPsi_Cp}
\begin{figure}[h!]
\centering
\begin{overpic}[height=0.417\linewidth,trim=0mm 5mm 0mm -6mm, abs]
               {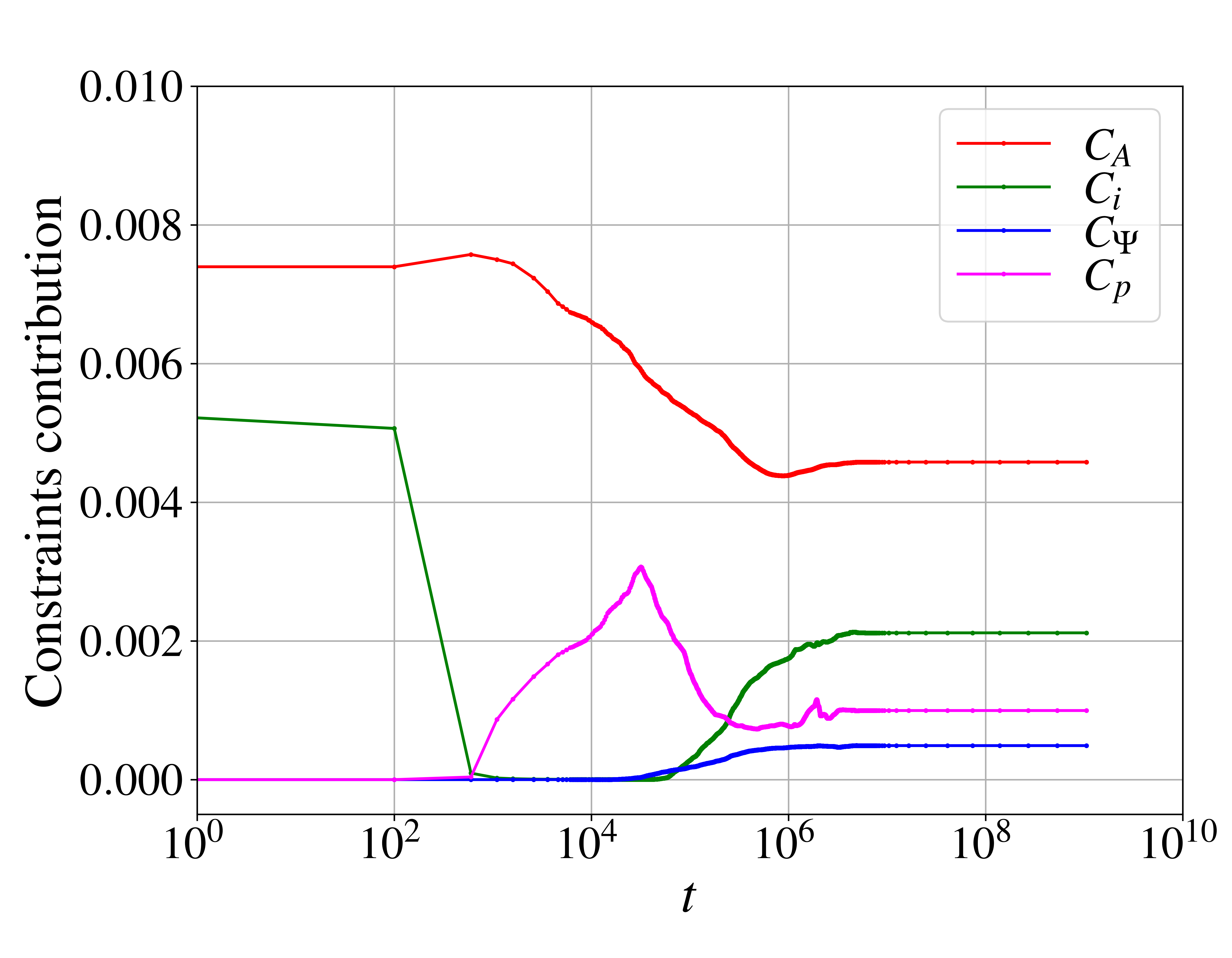}
\put(93,190){$\sqrt{2\Psi_\mathrm{lim}/E}=14\%$}
\end{overpic}
\begin{overpic}[height=0.417\linewidth,trim=31mm 5mm 0mm -6mm,clip, abs]
               {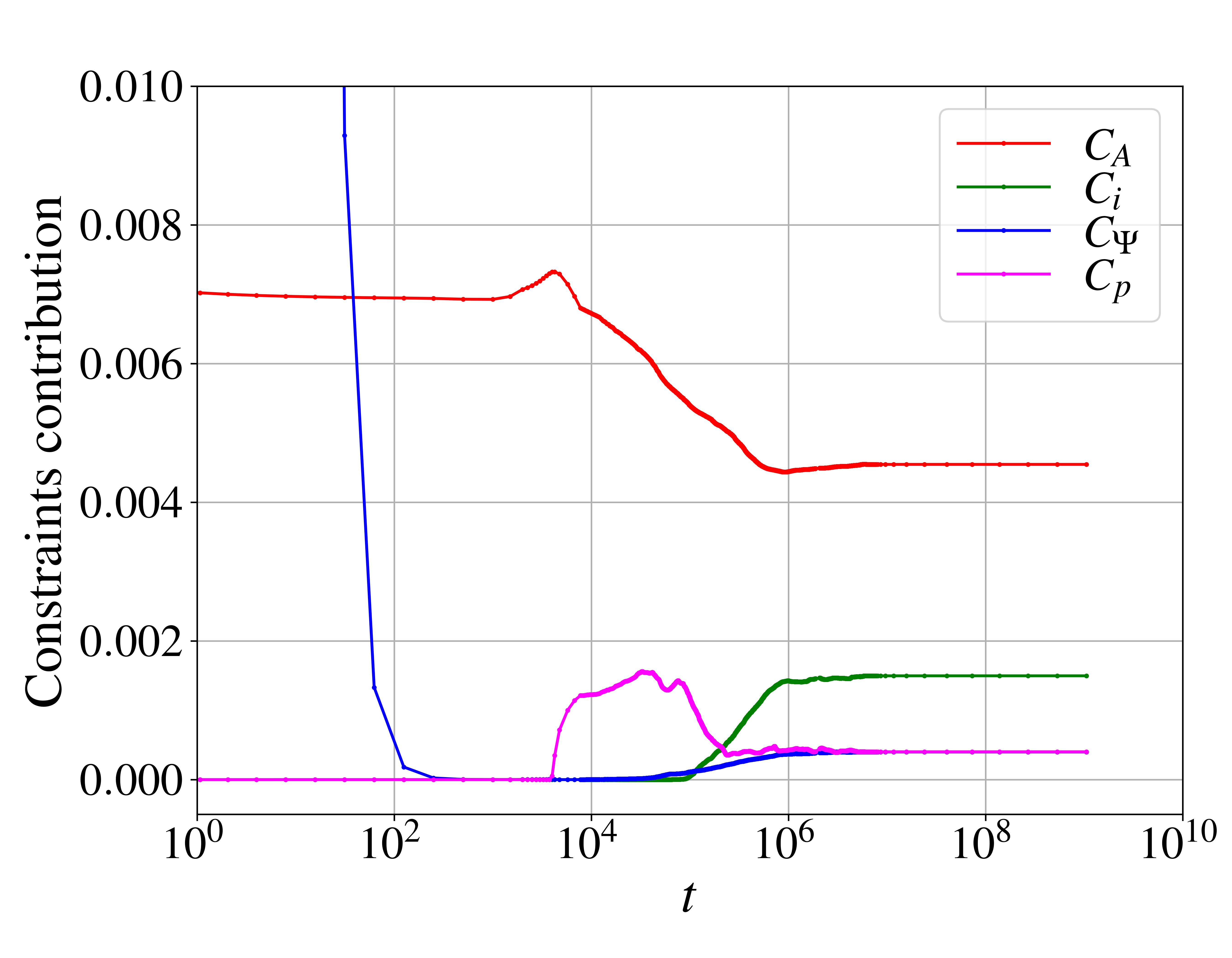}
\put(60,190){$\sqrt{2\Psi_\mathrm{lim}/E}=11\%$}
\put(-1,26){\textcolor{white}{\rule{5pt}{158pt}}}
\end{overpic}
\caption{Evolution of individual contributions $C_A$, $C_i$, $C_\Psi$, $C_p$, to the total objective.}
\label{fig:history_CA_Ci_CPsi_Cp}
\end{figure}

Figure~\ref{fig:history_CA_Ci_CPsi_Cp} shows the evolution of all penalization terms included in the objective function, during the optimization.
For the case with the stricter strain energy density limit constraint, this constraint is initially not satisfied.

\subsection{Spring calibration}
\label{appendix_spring_calib}
\begin{figure}[h!]
\centering
\begin{overpic}[width=0.9\textwidth,trim=19mm 72mm -59mm 75mm,clip]
               {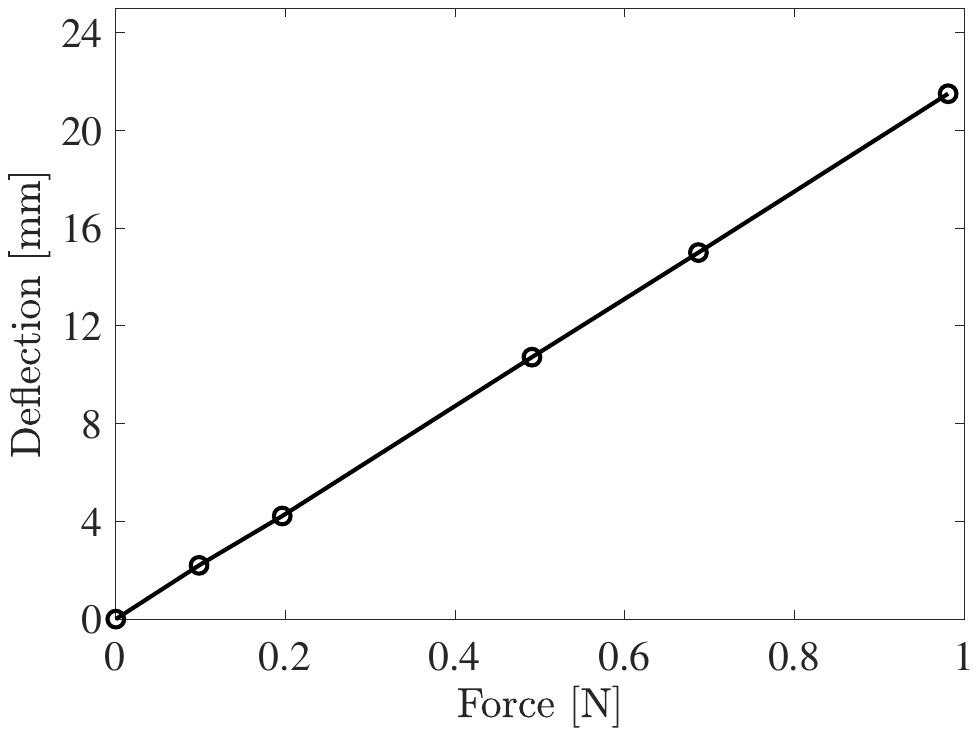}
  \put(68,2){\includegraphics[width=0.28\textwidth,trim=6cm 5cm 30cm 0cm,clip]
             {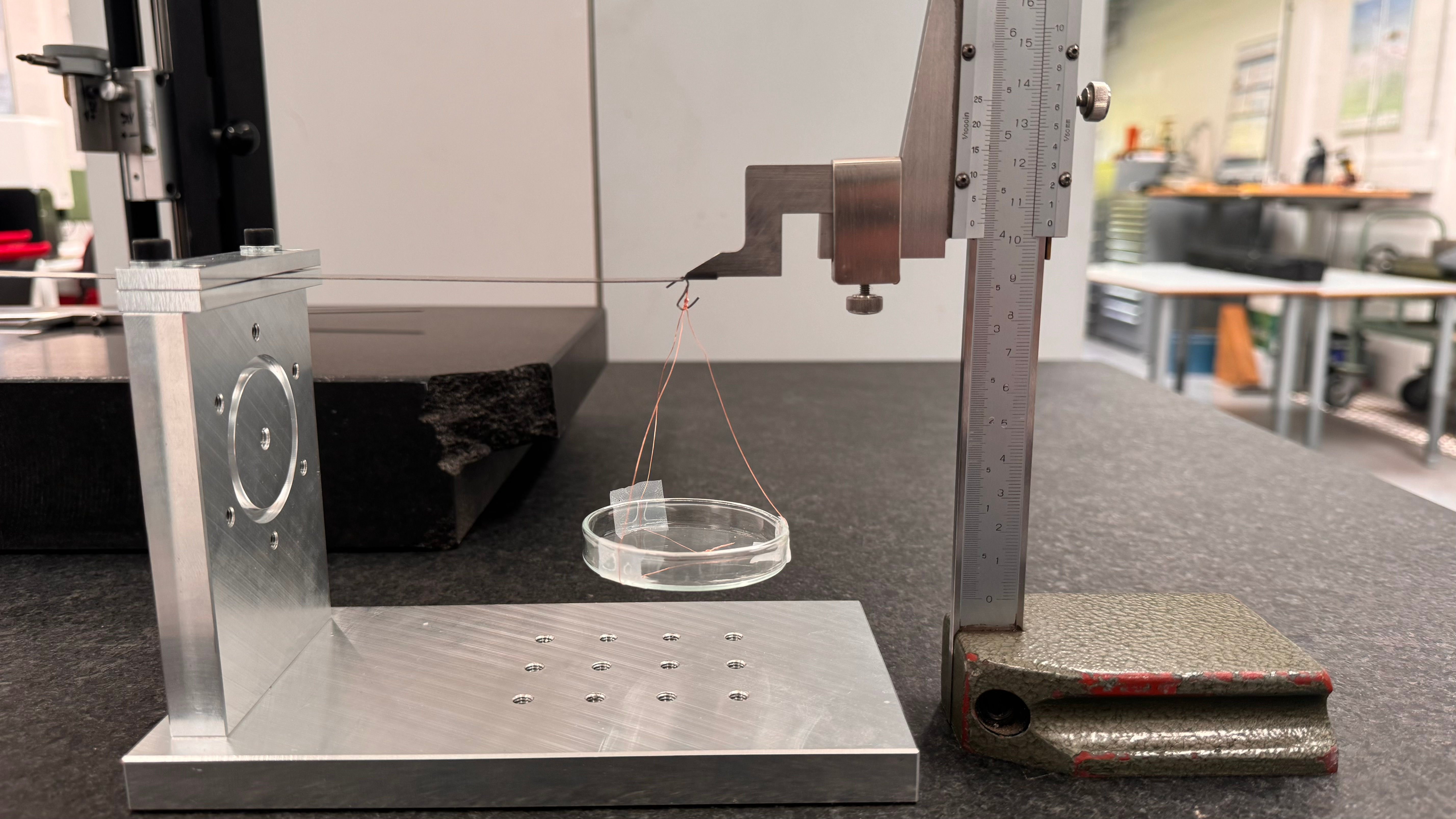}}
  \put(68,26){\includegraphics[width=0.28\textwidth,trim=6cm 5cm 30cm 0cm,clip]
              {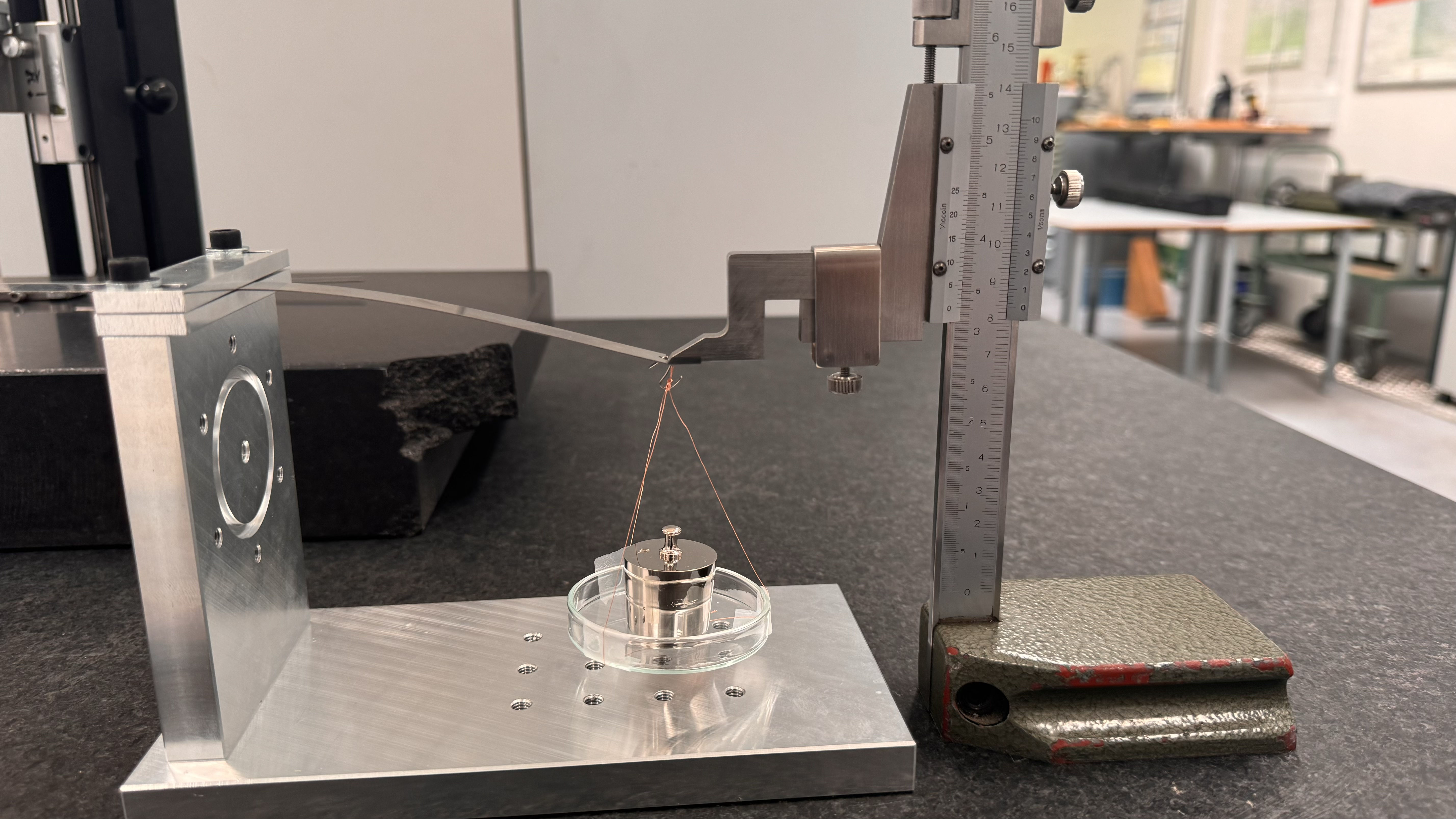}}
  \put(73,23){without weight}
  \put(73,47.5){with \SI{100}{\gram} weight}
\end{overpic}
\caption{Calibration of cantilever spring.}
\label{fig:cal_cantilever}
\end{figure}

Figure~\ref{fig:cal_cantilever} shows the results of the cantilever spring calibration as well as the calibration setup.
A height vernier gauge was used to measure deflections.
Weights corresponding to \SI{10}{\gram}, \SI{20}{\gram}, \SI{50}{\gram}, \SI{70}{\gram}, and \SI{100}{\gram} were used as loads.
The free length of the cantilever was calibrated at \SI{113}{\milli\meter} to reach the target stiffness of \SI{0.045}{\newton\per\milli\meter}, as shown in the force-deflection graph.

\subsection{Pressure and displacement measurements}
\label{appendix_raw_data}
\begin{figure}[h!]
\centering
\setlength{\tabcolsep}{0pt}
\begin{tabular}{c c}
$\sqrt{2\Psi_\mathrm{lim}/E}=14\%$ & $\sqrt{2\Psi_\mathrm{lim}/E}=11\%$\\
\includegraphics[width=0.49\linewidth,trim=0mm 5mm 0mm 12mm,clip]
                {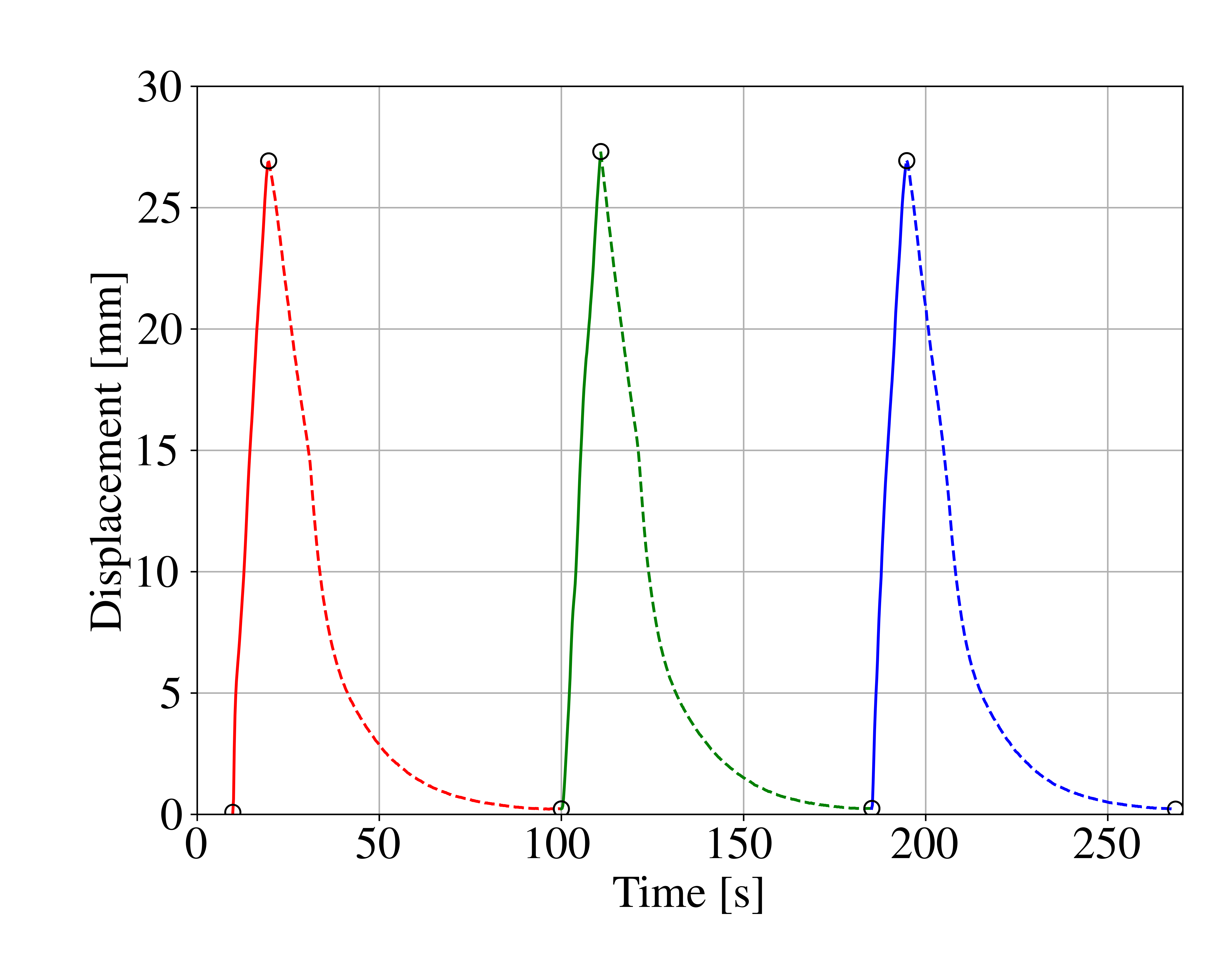}
&
\includegraphics[width=0.49\linewidth,trim=0mm 5mm 0mm 12mm,clip]
                {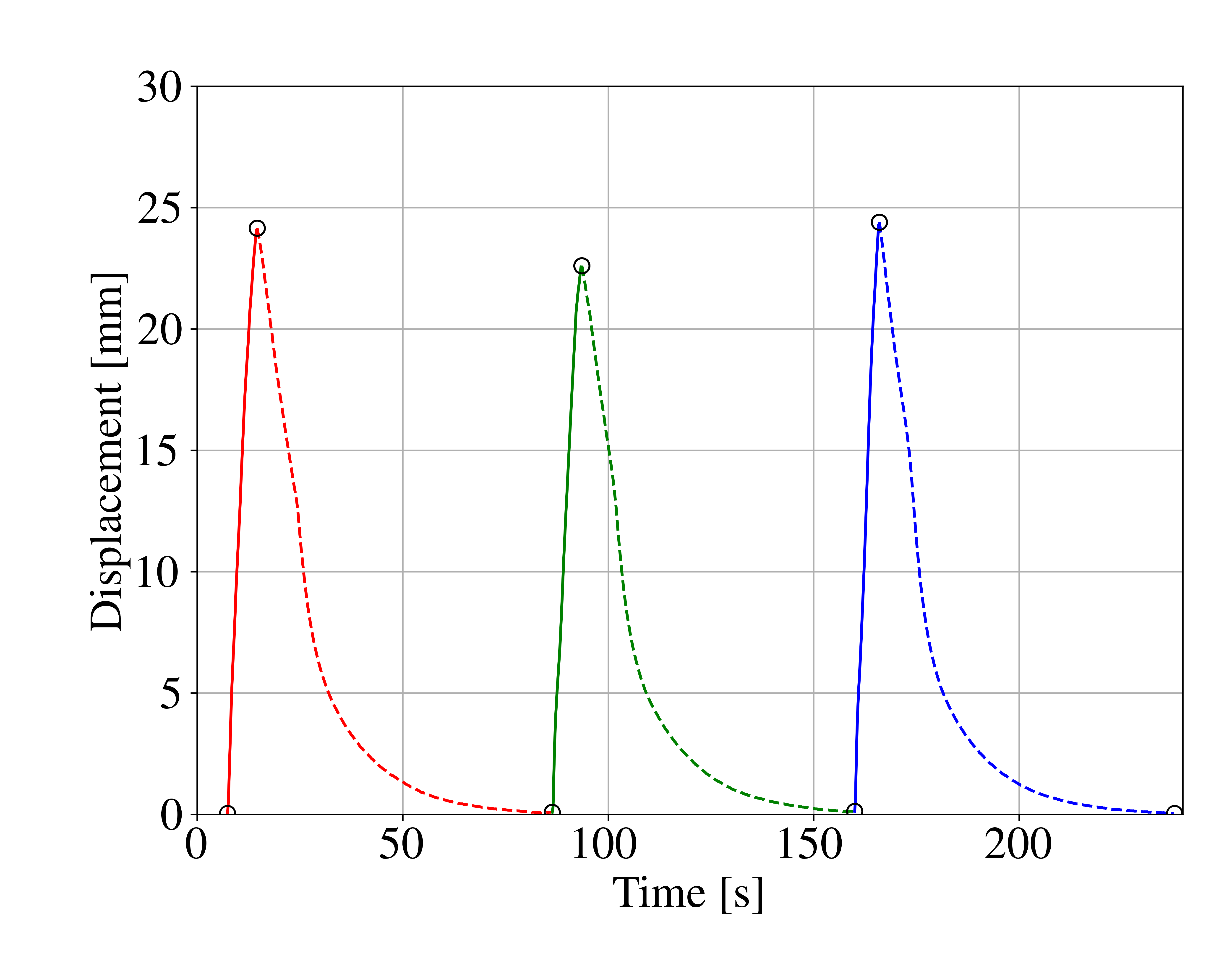}
\\
\includegraphics[width=0.49\linewidth,trim=0mm 5mm 0mm 12mm,clip]
                {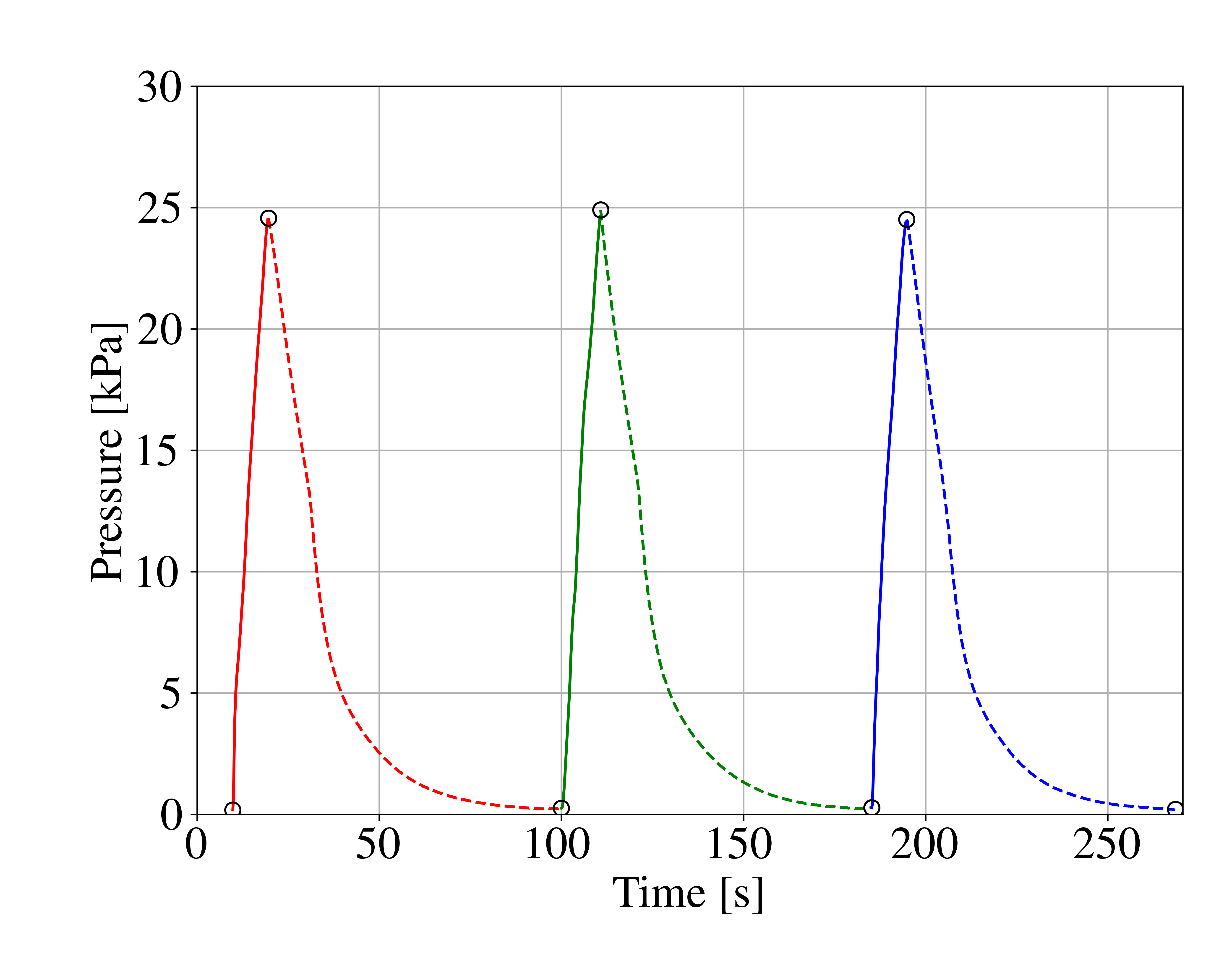}
&
\includegraphics[width=0.49\linewidth,trim=0mm 5mm 0mm 12mm,clip]
                {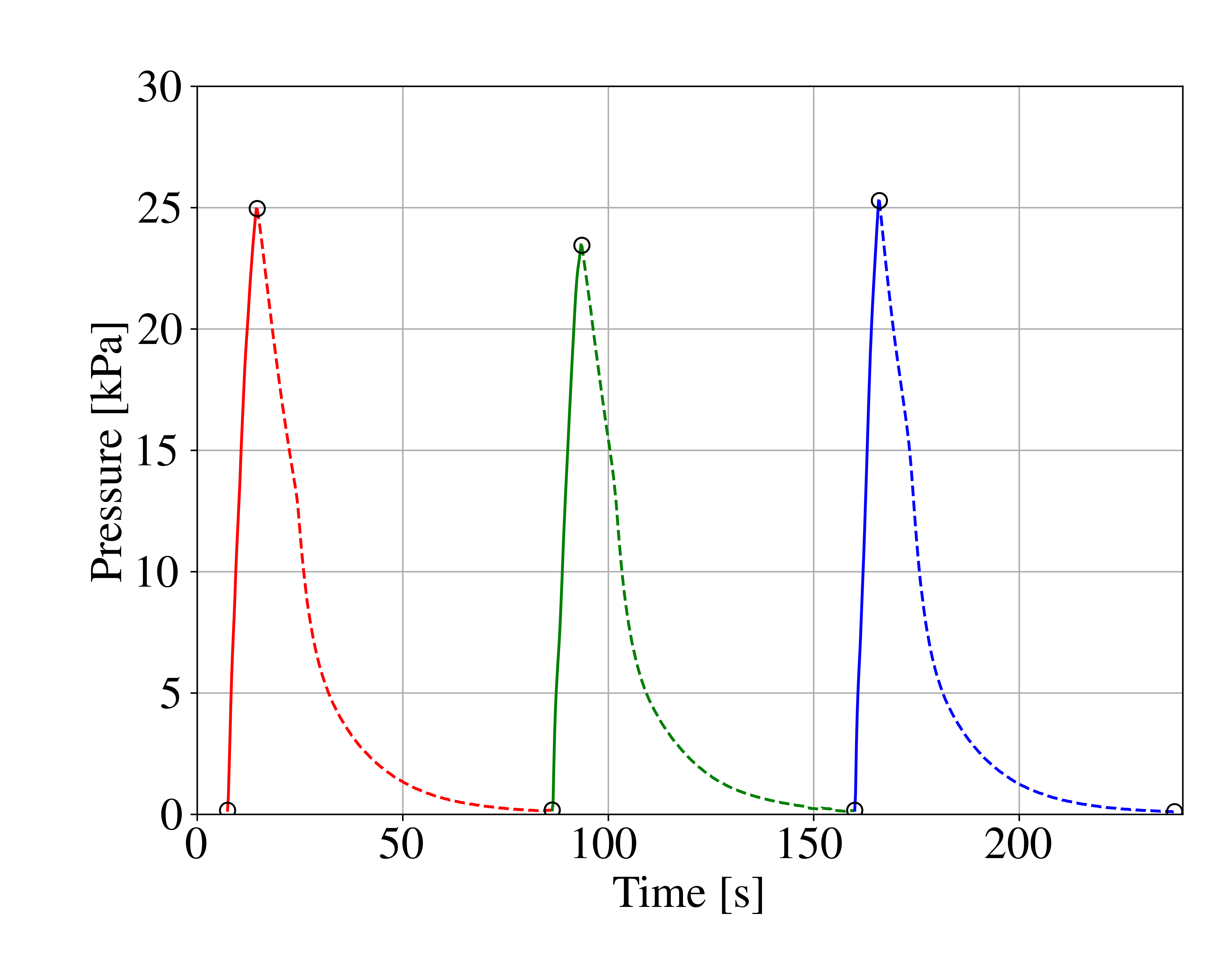}
\end{tabular}
\caption{Measured displacement and pressure data over time.}
\label{fig:raw_data}
\end{figure}

Figure~\ref{fig:raw_data} shows the original pressure and displacement data used in Figures~\ref{fig:exper_case14pct} and~\ref{fig:exper_case11pct}, as they were recorded over time.

\addcontentsline{toc}{section}{References}
\bibliographystyle{elsarticle-num}
\bibliography{references}

\end{document}